\documentclass[10pt,twocolumn,letterpaper]{article}

\usepackage{cvpr}
\usepackage[utf8]{inputenc}
\usepackage{times}
\usepackage{epsfig}
\usepackage{graphicx}
\usepackage{amsmath,mathtools}
\usepackage{amssymb}
\usepackage{amsthm}
\usepackage{nicefrac}       %
\usepackage{microtype}      %
\usepackage{amsmath} 
\usepackage{float}
\usepackage{amsfonts}
\usepackage{bm}
\usepackage{enumitem}
\usepackage{mathtools}
\usepackage{multirow}
\usepackage{fontawesome}
\usepackage{pifont}
\usepackage{color}
\usepackage{xcolor}
\usepackage[algo2e,inoutnumbered,linesnumbered,algoruled,vlined]{algorithm2e}
\usepackage{algpseudocode}
\usepackage{booktabs}
\usepackage{tabularx}
\usepackage{bm}
\usepackage{url}
\usepackage{bbm}
\usepackage{threeparttable}
\usepackage{tikz}
\usepackage{subfig}
\usepackage{wrapfig}
\usetikzlibrary{backgrounds}
\usetikzlibrary{decorations.pathreplacing}
\usepackage[outline]{contour}
\usepackage[pagebackref=true,breaklinks=true,colorlinks,bookmarks=false]{hyperref}
\usepackage[affil-it]{authblk}

\newif\ifHighResImages
\HighResImagestrue%

\newcommand{\parahead}[1]{\noindent\textbf{#1}:\ }

\newcommand{\LongName}{Neural Kernel Field\xspace}
\newcommand{\LongNames}{Neural Kernel Fields\xspace}
\newcommand{\ShortName}{NKF\xspace}
\newcommand{\ShortNames}{NKFs\xspace}

\theoremstyle{definition}

\newcommand{\RR}{\mathbb{R}}

\definecolor{darkblue}{RGB}{49,130,189}
\definecolor{stanfordgrey}{RGB}{46,45,41}
\definecolor{cardinalred}{RGB}{253,141,60}

\tikzset{>=stealth}
\tikzset{every node/.append style={text depth=0.7ex}}
\def\niterate{15}
\def\rolldice{
    \pgfmathsetmacro\rndlinewidth{9/(2+\i)}
    \pgfmathsetmacro\rndon{8+8*rnd}
    \pgfmathsetmacro\rndoff{2*rnd}
    \pgfmathsetmacro\rndshift{sqrt((1-\rndlinewidth/3)*14*rnd)}
    \pgfmathsetmacro\rndblend{98+\i*rand}
}
\tikzset{
    put dashes/.style={
        /utils/exec=\rolldice,
        line width=\rndlinewidth,
        dash pattern=on \rndon off \rndoff,
        dash phase=(\rndon+\rndoff)*rnd,
        shift={(rnd*360:\rndshift pt)},
        line cap=round,
        cardinalred,
        opacity=.8
    },
    chalk/.style={
        decorate,
        decoration={
            show path construction,
            lineto code={
                \foreach\i in{1,...,\niterate}{
                    \draw[put dashes]
                        (\tikzinputsegmentfirst)--(\tikzinputsegmentlast);
                }
            },
            curveto code={
                \foreach\i in{1,...,\niterate}{
                    \draw[put dashes]
                        (\tikzinputsegmentfirst)..controls
                        (\tikzinputsegmentsupporta)and(\tikzinputsegmentsupportb)
                        ..(\tikzinputsegmentlast);
                }
            },
            closepath code={
                \foreach\i in{1,...,\niterate}{
                \draw[put dashes]
                    (\tikzinputsegmentfirst)--(\tikzinputsegmentlast);
                }
            }
        }
    }
}
\tikzset{
    put blue_dashes/.style={
        /utils/exec=\rolldice,
        line width=\rndlinewidth,
        dash pattern=on \rndon off \rndoff,
        dash phase=(\rndon+\rndoff)*rnd,
        shift={(rnd*360:\rndshift pt)},
        line cap=round,
        darkblue,
        opacity=.8
    },
    chalk_blue/.style={
        decorate,
        decoration={
            show path construction,
            lineto code={
                \foreach\i in{1,...,\niterate}{
                    \draw[put blue_dashes]
                        (\tikzinputsegmentfirst)--(\tikzinputsegmentlast);
                }
            },
            curveto code={
                \foreach\i in{1,...,\niterate}{
                    \draw[put blue_dashes]
                        (\tikzinputsegmentfirst)..controls
                        (\tikzinputsegmentsupporta)and(\tikzinputsegmentsupportb)
                        ..(\tikzinputsegmentlast);
                }
            },
            closepath code={
                \foreach\i in{1,...,\niterate}{
                \draw[put blue_dashes]
                    (\tikzinputsegmentfirst)--(\tikzinputsegmentlast);
                }
            }
        }
    }
}

\cvprfinalcopy %

\def\cvprPaperID{4234} %

\ifcvprfinal\fi

\title{Neural Fields as Learnable Kernels for 3D Reconstruction}

\begin{document}

\renewcommand*{\Affilfont}{\normalsize\normalfont}
\renewcommand*{\Authfont}{\large\normalfont}
\renewcommand*{\Authsep}{, }
\renewcommand*{\Authand}{  }
\renewcommand*{\Authands}{ }
\makeatletter \renewcommand\AB@affilsepx{\hfill \protect\Affilfont} \makeatother
\newcommand\CoAuthorMark{\footnotemark[\arabic{footnote}]}
\author[1,2]{Francis Williams\thanks{Denotes equal contribution.}}
\author[1,3]{Zan Gojcic\protect\CoAuthorMark}
\author[1]{Sameh Khamis}
\author[2]{Denis Zorin}
\author[2]{\\\vspace{3mm}Joan Bruna}
\author[1,4,5]{Sanja Fidler}
\author[1]{Or~Litany}
\affil[1]{NVIDIA}
\affil[2]{New York University}
\affil[3]{ETH Z{\"u}rich}
\affil[4]{University of Toronto}
\affil[5]{Vector Institute}

\twocolumn[{%
\renewcommand\twocolumn[1][]{#1}%
\vspace{-20mm}
\maketitle
\begin{center}
\vspace{-5mm}
    \centering
    \captionsetup{type=figure}
    \begin{tikzpicture}
        \node (fig) {
    \ifHighResImages
        \includegraphics[width=0.95\textwidth]{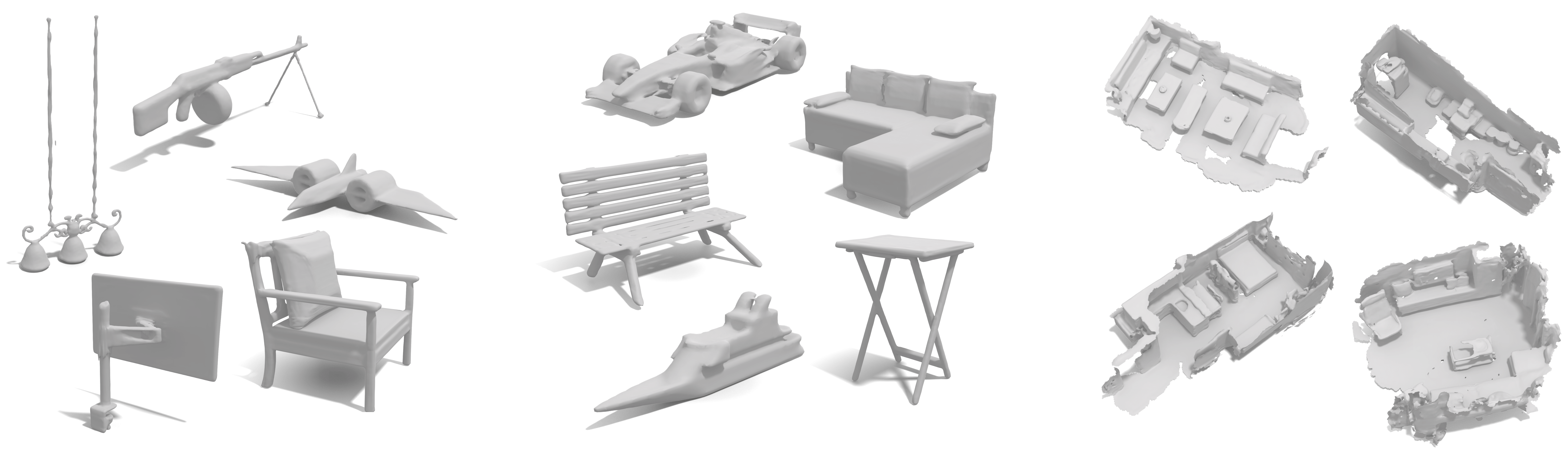}
    \else
        \includegraphics[width=0.95\textwidth]{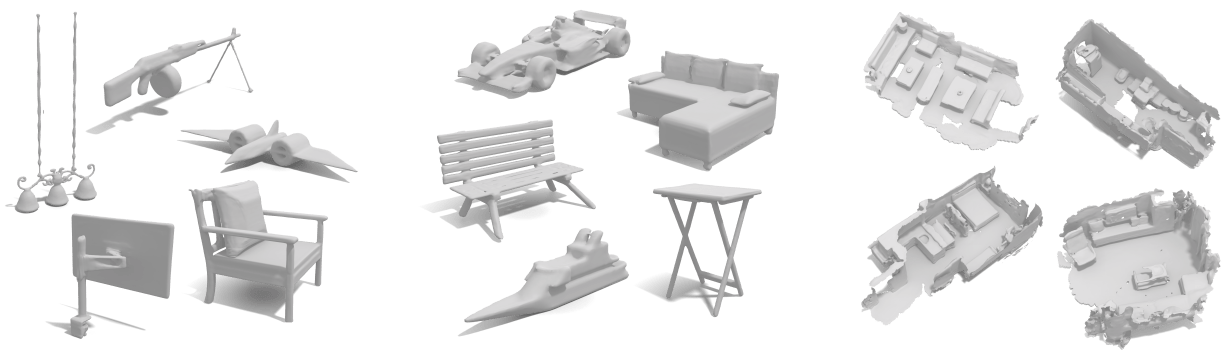}
    \fi};
    \node[anchor=west, text width=4.5cm] at (-7.75,-2.75) 
    {\small \textbf{In-category reconstruction}};
    \node[anchor=west, text width=4.5cm] at (-2.37,-2.75)
    {\small \textbf{Out-of-category reconstruction}};
    \node[anchor=west, text width=5cm] at (3.3,-2.75) 
    {\small \textbf{Generalization to scanned scenes}};
    \end{tikzpicture}
    \vspace{-2mm}
    \label{fig:teaser}\captionof{figure}{Trained on synthetic shapes, \ShortName  can reconstruct objects in and out of the training distribution, and scanned scenes.}
\end{center}%
\vspace{-1mm}
}]

\begin{abstract}
\vspace{-3mm}
We present \LongNames: a novel method for reconstructing implicit 3D shapes based on a learned kernel ridge regression. 
Our technique achieves state-of-the-art results when reconstructing 3D objects and large scenes from sparse oriented points, and can reconstruct shape categories outside the training set with almost no drop in accuracy. 
The core insight of our approach is that kernel methods are extremely effective for reconstructing shapes when the chosen kernel has an appropriate inductive bias. We thus factor the problem of shape reconstruction into two parts: (1) a backbone neural network which learns kernel parameters from data, and (2) a kernel ridge regression that fits the input points on-the-fly by solving a simple positive definite linear system using the learned kernel. As a result of this factorization, our reconstruction gains the benefits of data-driven methods under sparse point density while maintaining interpolatory behavior, which converges to the ground truth shape as input sampling density increases. Our experiments demonstrate a strong generalization capability to objects outside the train-set category and scanned scenes. 
Source code and pretrained models are available at \url{https://nv-tlabs.github.io/nkf}.

\end{abstract}

\vspace{-5mm}
\section{Introduction}\label{sec:introduction}
\vspace{-1mm}

The goal of 3D reconstruction is to recover geometry from partial measurements of a shape. In this work, we aim to map a sparse set of oriented points sampled from the surface of a shape to a 3D implicit surface for that shape. 
Surface reconstruction from point clouds is a well studied topic in computer vision and graphics, with applications in robotics, entertainment, and manufacturing. 
Techniques for surface reconstruction broadly fall into two types: implicit methods which aim to recover a volumetric function whose zero level-set encodes the surface, and explicit methods which directly recover a triangle mesh from the input points. While implicit approaches can adapt to arbitrary topologies, the requirement to store a dense volumetric field led many past works to favor explicit approaches~\cite{pan2019deep, gao2020learning}. More recently, implicit approaches have regained popularity due to a number of works demonstrating that neural networks are compact and effective at encoding signed-distance~\cite{park2019deepsdf, takikawa2021neural} and occupancy fields~\cite{mescheder2019occnet, peng2020convoccnet}. These works pair neural field\footnote{A \textit{neural field} refers to the parameterization of a continuous function of spatial coordinates using a neural network. In this work we focus on scalar functions mapping coordinates to real numbers.\\\hphantom{~~~~~}\textsuperscript{*}Denotes equal contribution.} representations with modern advances in point cloud processing architectures to produce powerful reconstruction techniques. Current state-of-the-art shape reconstructions methods can be categorized along three axes (Fig.~\ref{fig:related_work}):

\noindent
\textbf{(1) \textit{Feed-forward vs. test-time optimization}}: Feed-forward methods leverage shape priors to directly predict a surface from input points. While these methods are fast, they are not strictly constrained by their input and thus may perform a task more akin to retrieval than reconstruction (see~\cite{tatarchenko2019singleview} and Fig.~\ref{fig:motivation}, top). This results in decreased generalization performance on out-of-distribution shapes and input point densities. In contrast, test-time optimization via latent space traversal allows adaptation to the input, but is slow and can converge to poor local minima (See e.g.~\cite{duggal2021secrets} and Fig.~\ref{fig:motivation}, bottom). 

\noindent
\textbf{(2) \textit{Whether or not to leverage data priors}}: Data-free methods recover the surface by minimizing the residuals between the reconstructed surface and input points, leveraging a pre-determined prior to control the behavior away from the input points (e.g. a smooth space of functions~\cite{kazhdan2013screened, williams2020nsplines} or, emergent regularization arising from neural architectures~\cite{Williams_2019, gropp2020implicit}). Such fixed priors are, however, difficult to tailor to specific tasks, like completion of partial shapes (Fig.~\ref{fig:motivation}, middle). Data-driven approaches, on the other hand, can learn task-specific priors to predict shapes that resemble a given dataset.

\noindent
\textbf{(3) \textit{Which scale to process and represent data}}. 
Local-scale methods \cite{jiang2020lig, chabra2020deep} use the idea that complex structures can be reduced to a collection of simpler geometric primitives. These methods learn local models which are used to reconstruct a surface in patches. While this approach can generalize better, patch-size plays a critical role and must be carefully tuned per object (Fig.~\ref{fig:motivation}, bottom). Furthermore, without any notion of global context, these methods are unable to complete larger missing regions, leaving a fundamental gap in their generalization performance. 

Based on these axes and the motivating examples in Fig.~\ref{fig:motivation}, we identify the need for a method that can learn good priors from a simple collection of shapes to drive 3D reconstruction of both in-distribution and out-of distribution shapes and scenes. In particular, the priors learned by this method should respect the input points, performing reconstruction rather than retrieval.
\input{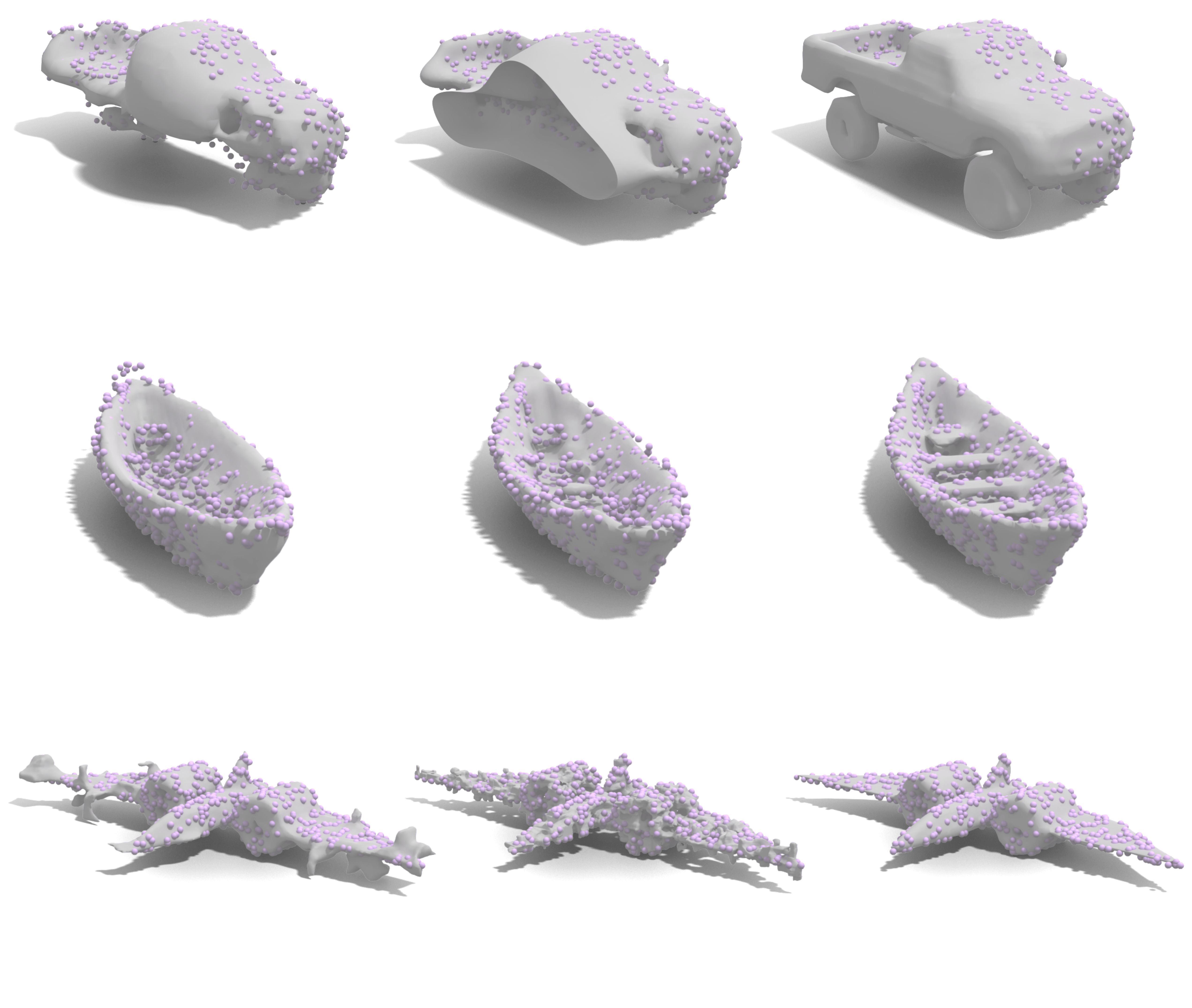}

We thus propose a method using a novel representation of neural fields based on learned kernels, which we call \emph{\LongNames} (\ShortNames).
In brief, \ShortNames work by learning a positive definite kernel conditioned on an input point cloud, and then using that kernel to predict an implicit shape by solving a simple linear system (Fig.~\ref{fig:architecture}).
Our approach provides several key benefits: First, since predicted kernels are conditioned on the input and learned from data, they enjoy the versatility of learning-based methods. Second, since \ShortNames leverage a kernel for shape prediction, any reconstructed surfaces respect the input points by construction. Third, unlike gradient descent-based latent space optimization, at test-time \ShortName kernel weights are solved in closed form via a simple convex least-squares problem, guaranteeing good minima. Finally, our kernel acts as a global aggregator of spatially local features, allowing our method to work at a wide variety of sampling densities without tuning any scale parameters.
The result is a generalizable method that can be trained only on synthetic shapes to seamlessly reconstruct out of distribution shapes and large scale scenes, while being robust to changes in input point density. Compared with the baselines, our method achieves a marked improvement reconstruction detail on both in and out-of distribution shapes. We summarize our contributions as follows:
\begin{itemize}
\setlength\itemsep{0.1em}
\item We introduce \LongNames, a novel representation of neural fields for 3D reconstruction, which outputs highly detailed surfaces that respect the input points. 
\item Our \ShortName representation achieves state of the art performance on ShapeNet reconstruction (Section~\ref{sec:reconstruction_shapenet}).
\item We show state-of-the-art generalization performance on out-of-distribution shapes (Section~\ref{sec:generalization}), scenes (Section~\ref{sec:reconstruction_scannet}) and point densities (Section~\ref{sec:convergence})
\end{itemize}

\section{Related Work}\label{sec:related}

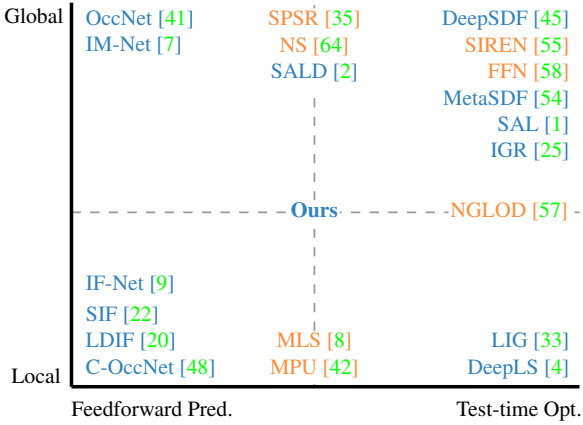
\begin{figure}
  \centering
  \resizebox{0.95\columnwidth}{!}{\begin{tikzpicture}
\draw[ line width=0.4 mm, color=black] (0.25,0.25) -- (6,0.25);
\draw[ line width=0.4 mm, color=black] (0.25,0.25) -- (0.25,4.6);
\draw[ line width=0.2 mm, color=black, dashed, opacity=0.4] (3,0.25) -- (3,4.6);
\draw[ line width=0.2 mm, color=black, dashed, opacity=0.4] (0.25,2.225) -- (6,2.225);
\node[ color=black, align=center,anchor=south,inner sep=0.0pt] at (-0.15,0.175) {\scriptsize Local};
\node[ color=black, align=center,anchor=north, inner sep=0.0pt] at (-0.175,4.55) {\scriptsize Global};
\node[ color=black, align=center,anchor=west] at (0.125,-0.05) {\scriptsize Feedforward Pred.};
\node[ color=black, align=center,anchor=east] at (6.15,-0.05) {\scriptsize Test-time Opt.};
\node [ color=darkblue, rotate=0, anchor=west, fill=white, rounded corners=2pt,inner sep=0.2pt] at (0.4,4.375) {\scriptsize OccNet~\cite{mescheder2019occnet}};
\node [ color=darkblue, rotate=0, anchor=west, fill=white, rounded corners=2pt,inner sep=0.2pt] at (0.4,4.075) {\scriptsize IM-Net~\cite{chen2019learning}};

\node [ color=darkblue, rotate=0, anchor=west, fill=white,rounded corners=2pt,inner sep=0.2pt] at (0.4,0.425) {\scriptsize C-OccNet~\cite{peng2020convoccnet}};
\node [ color=darkblue, rotate=0, anchor=west, fill=white, rounded corners=2pt,inner sep=0.2pt] at (0.4, 1.375) {\scriptsize IF-Net~\cite{chibane20ifnet}};
\node [ color=darkblue, rotate=0, anchor=west, fill=white,rounded corners=2pt,inner sep=0.2pt] at (0.4,1.025) {\scriptsize SIF~\cite{genova2019learning}};
\node [ color=darkblue, rotate=0, anchor=west, fill=white,rounded corners=2pt,inner sep=0.2pt] at (0.4,0.725) {\scriptsize LDIF~\cite{genova2020ldif}};

\node [ color=cardinalred, rotate=0, anchor=east, fill=white,rounded corners=2pt,inner sep=0.6pt] at (5.9,2.2) {\scriptsize NGLOD~\cite{takikawa2021neural}};
\node [ color=cardinalred, rotate=0, anchor=center, fill=white,rounded corners=2pt,inner sep=0.2pt] at (3.0,0.425) {\scriptsize MPU~\cite{ohtake2005multi}};
\node [ color=cardinalred, rotate=0, anchor=center, fill=white,rounded corners=2pt,inner sep=0.2pt] at (3.0,0.725) {\scriptsize MLS~\cite{cheng2008survey}};
\node [ color=darkblue, rotate=0, anchor=east, fill=white,rounded corners=2pt,inner sep=0.2pt] at (5.9,0.425) {\scriptsize DeepLS~\cite{chabra2020deep}};
\node [ color=darkblue, rotate=0, anchor=east, fill=white,rounded corners=2pt,inner sep=0.2pt] at (5.9,0.725) {\scriptsize LIG~\cite{jiang2020lig}};
\node [ color=cardinalred, rotate=0, anchor=center, fill=white,rounded corners=2pt,inner sep=0.6pt] at (3,4.375) {\scriptsize SPSR~\cite{kazhdan2013screened}};
\node [ color=cardinalred, rotate=0, anchor=center, fill=white,rounded corners=2pt,inner sep=0.6pt] at (3,4.075) {\scriptsize NS~\cite{williams2020nsplines}};
\node [ color=darkblue, rotate=0, anchor=center, fill=white,rounded corners=2pt,inner sep=0.6pt] at (3,3.775) {\scriptsize SALD~\cite{atzmon2020sald}};
\node [ color=darkblue, rotate=0, anchor=east, fill=white,rounded corners=2pt,inner sep=0.2pt] at (5.9,4.375) {\scriptsize DeepSDF~\cite{park2019deepsdf}};
\node [ color=cardinalred, rotate=0, anchor=east, fill=white,rounded corners=2pt,inner sep=0.2pt] at (5.9,4.075) {\scriptsize SIREN~\cite{sitzmann2020implicit}};
\node [ color=cardinalred, rotate=0, anchor=east, fill=white,rounded corners=2pt,inner sep=0.2pt] at (5.9,3.775) {\scriptsize FFN~\cite{tancik2020fourfeat}};
\node [ color=darkblue, rotate=0, anchor=east, fill=white,rounded corners=2pt,inner sep=0.2pt] at (5.9,3.475) {\scriptsize MetaSDF~\cite{sitzmann2020metasdf}};
\node [ color=darkblue, rotate=0, anchor=east, fill=white,rounded corners=2pt,inner sep=0.2pt] at (5.9,3.175) {\scriptsize SAL~\cite{atzmon2019sal}};
\node [ color=darkblue, rotate=0, anchor=east, fill=white,rounded corners=2pt,inner sep=0.2pt] at (5.9,2.875) {\scriptsize IGR~\cite{gropp2020implicit}};
\node [ color=darkblue, rotate=0, anchor=center, fill=white,rounded corners=2pt,inner sep=0.6pt] at (3,2.2) {\scriptsize \textbf{Ours}};

\end{tikzpicture}}
  \vspace{-4mm}
  \caption{Taxonomy of design choices for methods which reconstruct implicit shapes from point clouds. The $x$ and $y$ axes correspond to axes 1 and 3 discussed in Section~\ref{sec:introduction} respectively. Color corresponds to axis 2:  {\color{darkblue}{\textbf{blue} methods use learned priors}}, and {\color{cardinalred}{\textbf{orange} methods do \emph{not}}}.}
  \label{fig:related_work}
  \vspace{-4mm}
\end{figure}

Figure~\ref{fig:related_work} visualizes existing implicit 3D shape reconstruction methods along the three axes defined in Section~\ref{sec:introduction}. Our \LongName approach lies at the center of the diagram since it (1) uses a simple \textit{convex} test time optimization, (2) leverages priors learned from data, and (3) learns local features on a spatial grid, but aggregates these globally during fitting. 

We now highlight several works that are particularly relevant to our approach: Learned kernels were investigated in~\cite{wilson2015deep, NIPS2007_4b6538a4, patacchiola2020bayesian} and used for tasks such as few-shot transfer learning and classification of images. Neural Splines~\cite{williams2020nsplines} used a kernel method derived from infinitely wide ReLU networks to reconstruct 3D surfaces from points. Convolutional Occupancy Networks~\cite{peng2020convoccnet} proposes a convolutional architecture that maps 3D points to features. We use a similar feature network for our \LongName architecture. LIG~\cite{genova2020local} addresses the need for reconstruction methods that can generalize. MetaSDF~\cite{sitzmann2020metasdf} meta-learns a network which can be rapidly trained to predict SDFs. \LongNames can also be viewed as a form of meta-learning since they predict a kernel machine from data. Shape as Points~\cite{peng2021shape} is a concurrent work relevant to our method. It solves a linear system to reconstruct a surface after a learned upsampling phase. Unlike our method, however, Shape as Points relies on the inductive bias of Poisson reconstruction to output a surface rather than learning an inductive bias from data.

Beyond methods based on implicit surfaces, other shape reconstruction techniques exist which leverage different output representations. These representations include dense point clouds \cite{rempe2020caspr, luo2021diffusion, zhou20213d, qi2017pointnet, qi2017pointnetpp, zhao20193d, sun2020canonical, yu2018ecnet, yu2018punet,fan2016point,lin2017learning}, polygonal meshes \cite{hanocka2020point2mesh, chen2020bspnet, gao2020learning, Hanocka_2019, gkioxari2020mesh, Williams_2020_CVPR_Workshops, deng2020cvxnet,Litany_2018_CVPR,halimi2020greater,dmtet21}, manifold atlases \cite{Williams_2019, deprelle2019learning, groueix2018atlasnet, gadelha2020deep, badki2020meshlet}, and voxel grids \cite{choy20163dr2n2,tulsiani2020objectcentric,hane2017hierarchical,marrnet,tulsiani2017multiview,girdhar2016learning}. While our method focuses on shape reconstruction from points, past work has used neural fields to perform a variety of 3D tasks such as shape compression~\cite{takikawa2021neural, williams2020nsplines}, shape prediction from images~\cite{mescheder2019occnet, lin2020sdfsrn}, voxel grid upsampling~\cite{peng2020convoccnet, mescheder2019occnet}, reconstruction from rotated inputs~\cite{deng2021vector} and articulated poses~\cite{deng2020nasa,zhang2021strobenet}, and video to 3D~\cite{yariv2020multiview, liu2020dist}.

\section{Method}\label{sec:method}
\begin{figure*}
    \centering
    \ifHighResImages
        \includegraphics[width=0.94\textwidth]{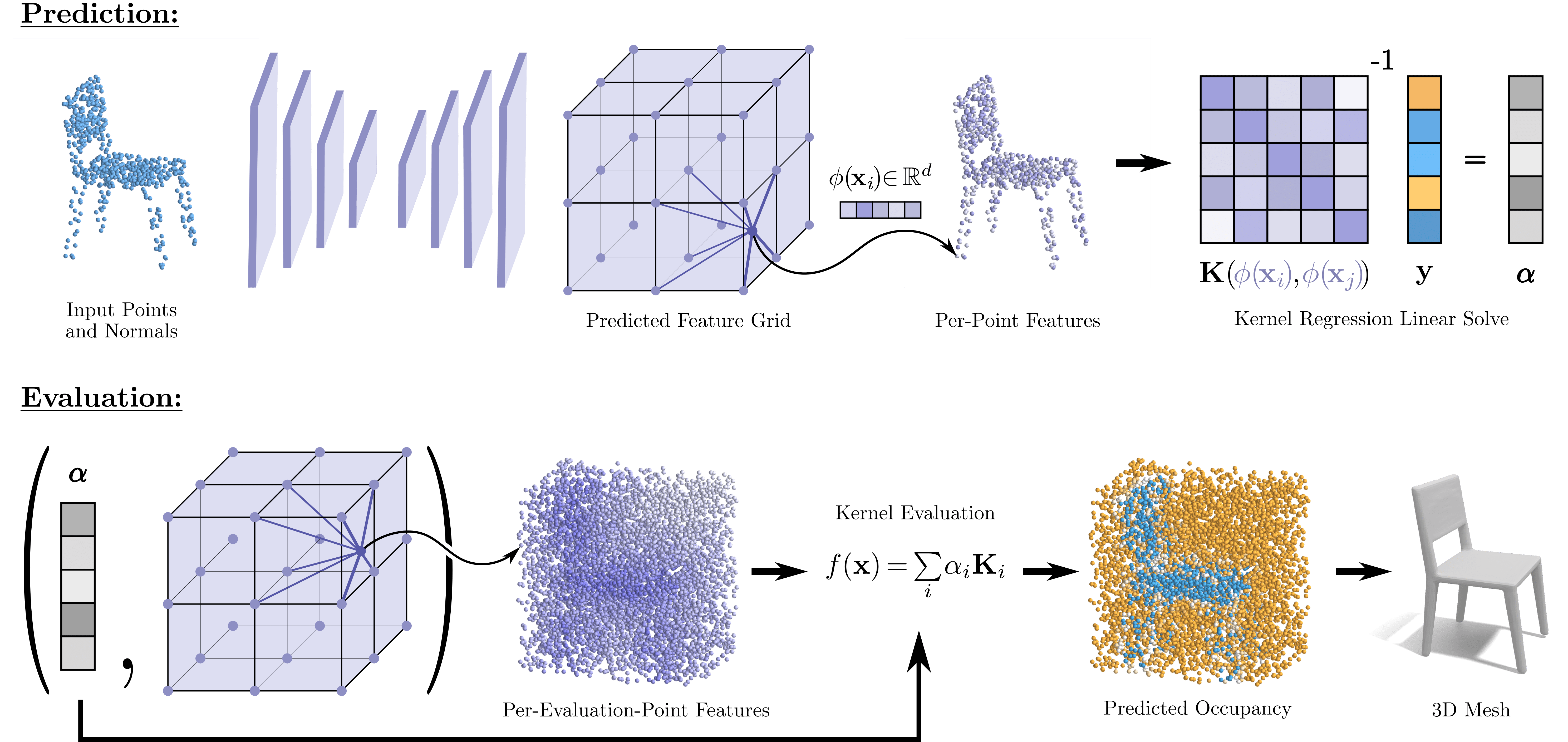}
    \else
        \includegraphics[width=0.94\textwidth]{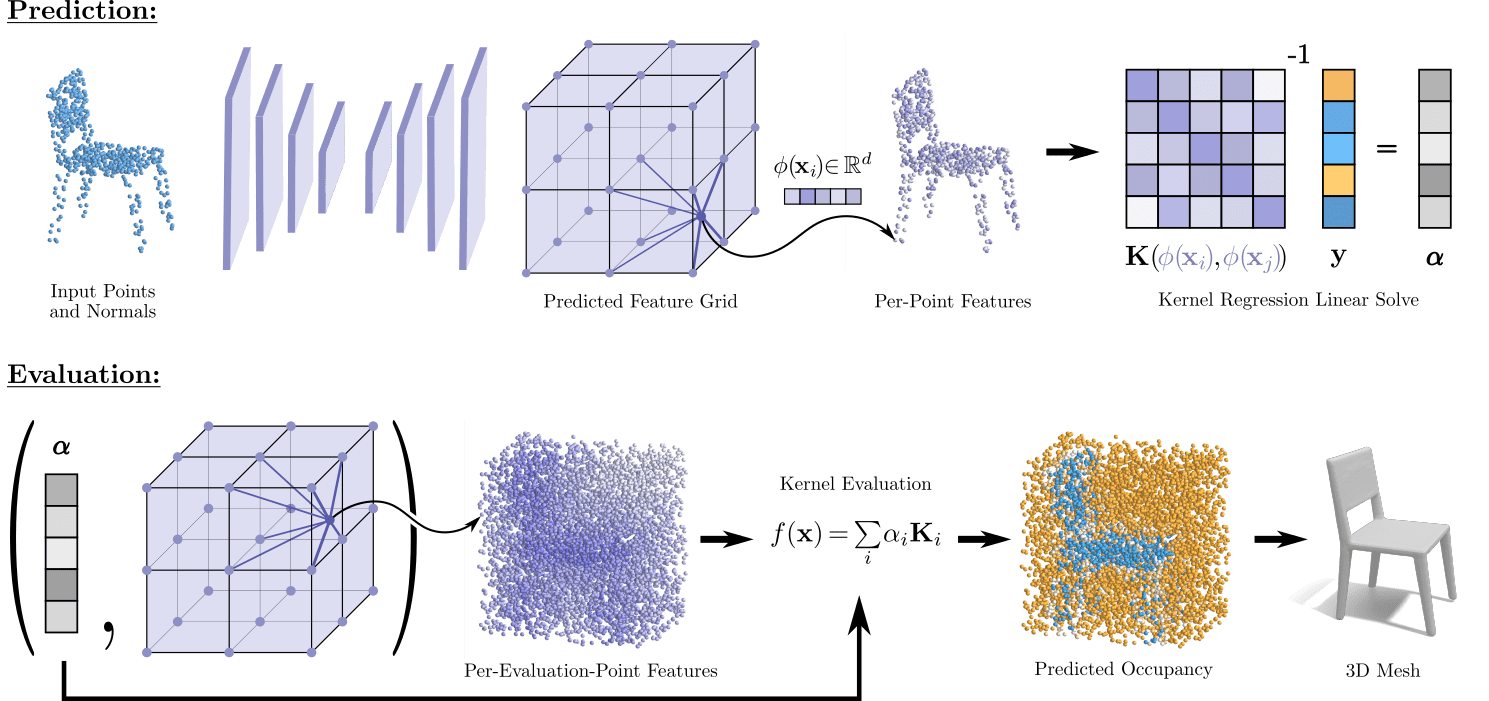}
    \fi
        
    \caption{Our method works in two stages: (1) \emph{prediction} (Top row) where we predict an implicit function from an input point cloud, and (2) \emph{evaluation} (Bottom row) where we evaluate the implicit function. Our predicted implicit consists of a feature function $\phi$ which lifts points in the volume to features in $\RR^d$, and a set of coefficients $\bm \alpha$, which are used to encode the function as a linear combination of basis functions centered at the input points.}
    
    \label{fig:architecture}
     \vspace{-4mm}
\end{figure*}

Our approach predicts an implicit surface from an oriented point cloud using a learned kernel. Neural Splines~\cite{williams2020nsplines} also solves a 3D reconstruction problem using a fixed kernel (not learned from data), and is thus related to our approach. To introduce the reader to kernel methods for 3D reconstruction, we begin by giving an overview of Neural Splines. We then show how these kernel methods can be extended into \LongNames capable of leveraging priors from data.

\subsection{Review of Neural Splines}\label{sec:neural_splines_review}
Given a point set $X = \{\bm x_i \in \RR^3\}_{i=1}^S$ with corresponding normals $N = \{\bm n_i \in \RR^3\}_{i=1}^S$, \cite{williams2020nsplines} seeks an implicit field $f : \RR^3 \rightarrow \RR$ which represents the underlying surface from which $X$ and $N$ were sampled. Namely, it should zero out on the set of input points and its gradient should equal the normal direction. More formally, the implicit field should minimize
\begin{equation}\label{eq:ns_loss}
    L(f) = \sum_{i=1}^S |f(\bm x_i)|^2 + \|\nabla f(\bm x_i) - \bm n_i\|^2    
\end{equation}
\begin{wrapfigure}{r}{0.3\columnwidth}
    \centering
    \includegraphics[width=0.3\columnwidth]{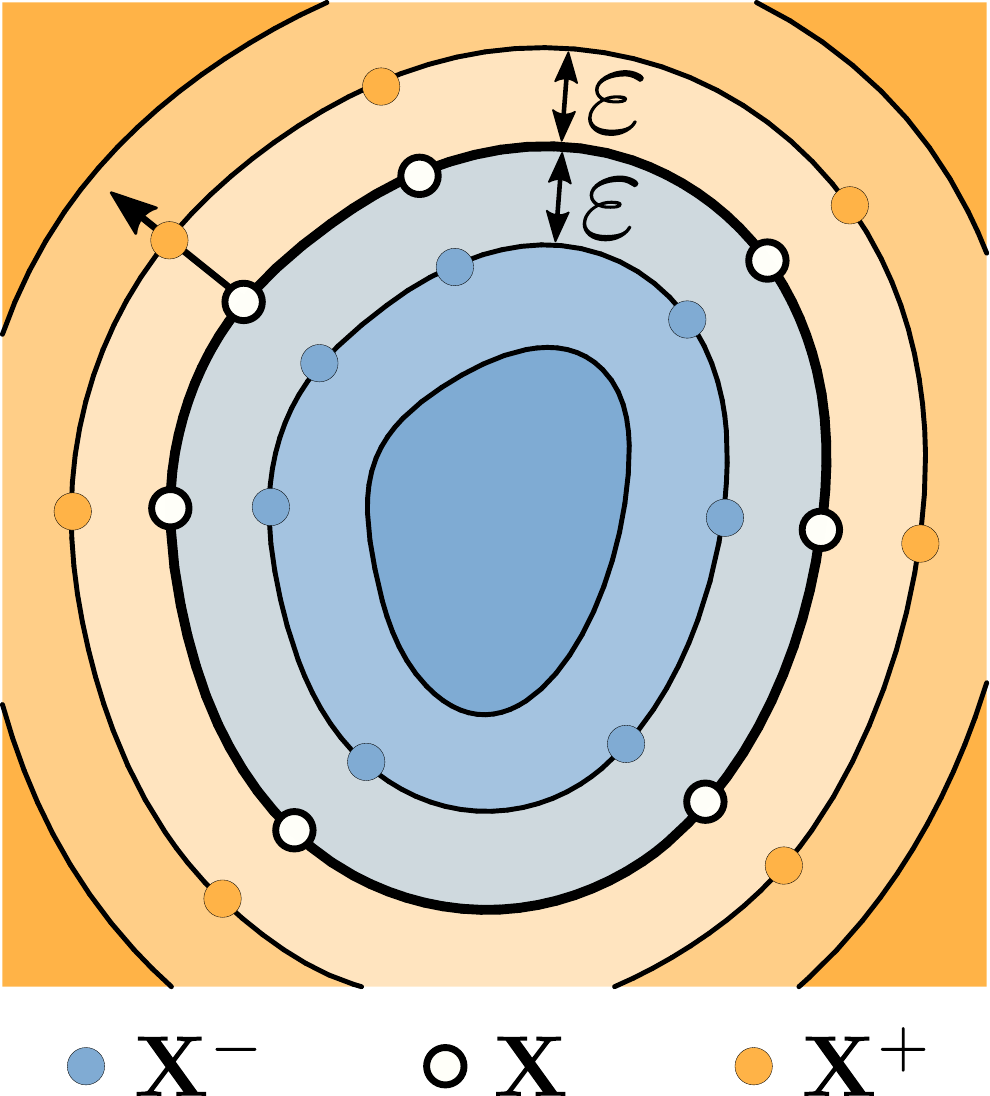}
    \label{fig:tripled_points}
    \vspace{-4em}
\end{wrapfigure}

The gradient part of \eqref{eq:ns_loss} can be approximated with a finite difference method, by augmenting the points $X$ with $X^+ = \{\bm x_i^{+} = \bm x_i + \epsilon \bm n_i\}_{i=1}^S$ and $X^- = \{\bm x_i^{-} = \bm x_i - \epsilon \bm n_i\}_{i=1}^S$ (see inset figure)
and minimizing the simpler loss:
\begin{equation}\label{eq:simple_loss}
    L(f) = \sum_{i=1}^S |f(\bm x_i)|^2 + | f(\bm x_i^{+}) - \epsilon|^2 + | f(\bm x_i^{-}) + \epsilon|^2
\end{equation}

Let $X' = X \cup X^+ \cup X^-$ denote the union of the augmented points. To minimize \eqref{eq:simple_loss}, we represent $f$ as a weighted sum of kernel basis functions centered at the points $X'$:
\begin{equation}
    f(\bm x) = \sum_{\bm x' \in X'} \alpha_i K_\text{NS}(\bm x, \bm x')
\end{equation}
which is linear in the coefficients $\bm \alpha = \begin{bmatrix} \alpha_1 & \ldots & \alpha_{3S} \end{bmatrix}^T$. These coefficients can thus be recovered by solving the linear system
\begin{equation}\label{eq:linsys}
(\bm G + \lambda \bm I) {\bm \alpha} = \bm y
\end{equation}
where $\bm G \in \RR^{3S \times 3S}$ is the augmented Gram matrix over the points $X'$ (\textit{i.e.} $\bm G_{ij} = K_\text{NS}(\bm x'_i, \bm x'_j) \: \forall \bm x_i', \bm x_j' \in X'$), $\lambda > 0$ is an optional regularizer which can be used to filter noise, and $\bm y$ is a vector such that
\begin{equation}
    y_j = \begin{cases} 
    0         &\text{if } \bm x_j' \in X\\
    +\epsilon &\text{if } \bm x_j' \in X^+\\
    -\epsilon &\text{if } \bm x_j' \in X^-
    \end{cases}
\end{equation}

The kernel function $K_\text{NS}$ is the closed form expression for an infinitely wide shallow ReLU network. It depends on the inner product between the inputs expressed in homogeneous coordinates. i.e. $K_\text{NS}(x, y) = K_\text{NS}(\langle x, y \rangle + 1)$. See the appendix for the exact equation and more details.%

\subsection{Inductive Bias of Neural Splines}
The kernel formulation in Neural Splines makes explicit the notion of \emph{inductive bias}, \textit{i.e.} the behavior of solutions away from the input points. To see this, we observe that solutions to the linear system \eqref{eq:linsys} are solutions to the following constrained optimization problem:
\begin{gather}
    \text{minimize } \|f\|_K = \bm \alpha^T \bm G \bm \alpha \\
    \text{subject to } f(\bm x'_j) = y_j \qquad \bm x'_j \in X'
\end{gather}
Here the norm $\|f\|_K$ being minimized defines the inductive bias of the kernel method, \textit{i.e.} it governs the behavior of the function away from the constraints. The constraints $f(\bm x_j) = y_j$ guarantee that any solution to the above optimization problem interpolates the input data up to a bound defined by the regularizer $\lambda$. 

For Neural Splines, the kernel norm favors smooth functions: It is proportional to curvature ($\|f\|_K \approx \|f''\|$) for 1D curves \cite{williams2019gradient} and to the Radon transform of the Laplacian ($\|f\|_K \approx \|\mathcal{R}\{\Delta^2 f\}\|$) for 3D implicit surfaces \cite{ongie2019function, williams2020nsplines}. 
While an inductive bias favoring smoothness is good for reconstructing shapes with dense samples, it is too weak a prior in more challenging cases such as when the input points are very sparse or only cover part of a shape. For example, Fig.~\ref{fig:motivation} (top) shows that Neural Splines is incapable of completing a partial point cloud of a truck. To this end, \ShortNames use a \emph{data dependent kernel}, which learns an appropriate inductive bias conditioned on the input. By solving a linear system such as \eqref{eq:linsys} using this kernel, we guarantee that output shapes respect their input points. We now describe \ShortNames in detail. 

\vspace{-0.25em}
\subsection{\LongNames}\label{sec:neural_kernel_fields}\vspace{-0.5em}
Our model accepts the same inputs as Neural Splines described above in Section~\ref{sec:neural_splines_review}: \textit{i.e.} We are given a set of points $X$ and normals $N$ sampled from the surface of an unknown shape, which we subsequently expand into an augmented point cloud $X'$ with $2S$ points and corresponding labels $\bm y \in \RR^{2S}$. We remark that our method only uses the inside and outside augmented points, \textit{i.e.} $X' = X^+ \cup X^-$. For brevity, we denote the inputs to our model as $\mathcal{X} = (X', \bm y)$. We now describe our architecture in four steps: (1) how to define our data dependent kernel, (2) how to use that kernel to predict an implicit function, (3) how to train our model, and (4) how to add filtering for noisy inputs. Figure~\ref{fig:architecture} shows our \ShortName architecture pictorially.

\vspace{-1.0em}
\paragraph{Data Dependent Kernel} To learn a kernel from data, we first augment input points $\bm x'_i \in X'$ with a feature $\phi(\bm x'_i | \mathcal{X}, \theta) \in \RR^d$ where $\phi$ is a neural network with parameters $\theta$ conditioned on the inputs $\mathcal{X}$. Using these learned per-point features, we the define data-dependent kernel as:
\begin{equation}\label{eq:data_kernel}
    K_{(\mathcal{X}, \theta)}(\bm x, \bm z) =
    K_{\text{NS}}([\bm x : \phi(\bm x | \mathcal{X}, \theta)], [\bm z : \phi(\bm z | \mathcal{X}, \theta)])
\end{equation}
where $[\bm a : \bm b]$ is the concatenation of the vectors $\bm a$ and $\bm b$, and $K_{NS}$ is the Neural Spline kernel function. 
The architecture of the network $\phi$ follows an approach similar to Convolutional Occupancy Networks \cite{peng2020convoccnet}: We discretize the volume around the input point cloud into a $M\times M \times M$ grid, and use a PointNet within each grid cell containing input points to extract a feature in that cell (empty cells have a zero feature). We then feed these features into a fully convolutional 3D U-Net, which produces an $M \times M \times M \times d$ grid of output features. To extract features per point, we trilinearly interpolate the output grid using the sampled points.

\vspace{-1em}
\paragraph{Predicting an Implicit Function}
To predict an implicit function, we find coefficients $\alpha_j$ for each input point $\bm x'_j \in X'$ by solving the $2S \times 2S$ positive definite linear system
\begin{equation}\label{eq:linsys_learned}
    \bm \alpha = [\alpha_j]_{j=1}^{2S} = (\bm G(\mathcal{X}, \theta) + \lambda \bm I)^{-1} \bm y
\end{equation} where $\bm G(\mathcal{X}, \theta)$ is the gram matrix $\bm G(\mathcal{X}, \theta)_{ij} = K_{(\mathcal{X}, \theta)}(\bm x_i, \bm x_j)$, and $\lambda$ is a user supplied regularization parameter. To evaluate the predicted function at a new point $\bm x$, we compute the following equation using the coefficients $\bm \alpha$:
\begin{equation}
    f(\bm x) = \sum_{\bm x'_j \in X'} \alpha_j K_{(\mathcal{X}, \theta)}(\bm x, \bm x'_j).
\end{equation}

\vspace{-1.5em}
\paragraph{Training the Model} To supervise our model during training, we use a dataset of shapes. Each shape consists of the augmented input points and labels $\mathcal{X} = (X', \bm y)$, a dense set of points and occupancy labels ($X_\text{vol} = \{\bm x_i^\text{vol} \in \RR^3\},  Y_\text{vol} = \{y_i^\text{vol} \in \RR\}$) in the volume surrounding the shape, and a dense set of points $X_\text{surf} = \{\bm x_i^\text{surf} \in \RR^3\}$ on the surface of the shape. We remark that the dense points on the surface and in the volume are only needed as supervision during training. The occupancy labels $Y_\text{vol}$ denote whether a volume point lies inside or outside a shape and are defined as: 
\begin{equation}
    y_i^\text{vol} = \begin{cases} 1 & \text{if } \bm x_i^\text{vol} \text{ is inside the shape}\\ 0 & \text{otherwise}\end{cases}
\end{equation}
We then train the network $\phi$ used to define the kernel \eqref{eq:data_kernel} by first predicting an implicit function using the inputs $\mathcal{X}$ and then evaluating it at the dense volume $X_\text{vol}$ and surface $X_\text{surf}$ points to compute the loss:
\begin{equation}\label{eq:supervision_loss}
    L(f) = \sum_{i = 1}^{|X_\text{vol}|}\text{BCE}(f(\bm x_i^\text{vol}), y_i^\text{vol}) + \lambda_{L1} \sum_{i=1}^{|X_\text{surf}|}|f(\bm x_i^\text{surf})|
\end{equation}
The first term in \eqref{eq:supervision_loss} encourages the predicted function to have the correct occupancy, while the second term encourages the surface to agree with the ground truth shape. We backpropagate gradients through this loss to update the weights of the network $\phi$, and thus learn the data dependent kernel. 

\begin{figure}
    \minipage{0.5\columnwidth}
    \centering
    \includegraphics[width=0.97\columnwidth]{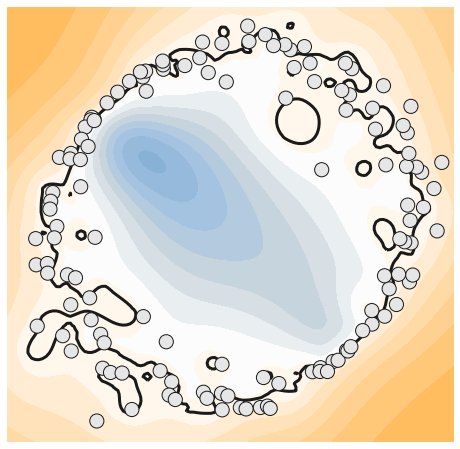}
    \endminipage
    \centering
    \minipage{0.5\columnwidth}
    \centering
    \includegraphics[width=0.97\columnwidth]{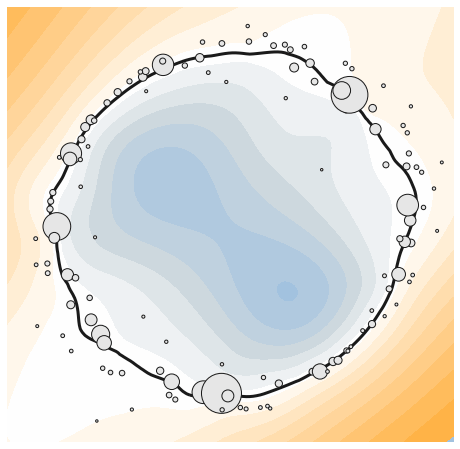}
    \endminipage
    \caption{Unweighted (left) versus weighted (right) kernel ridge regression. Both reconstructions use the same noisy input points and regularization value. The right reconstruction, which uses per-point weights (visualized as the size of the points) can filter out the contribution of noisy points and produce a more accurate reconstruction.}
    \label{fig:weighted_ridge_regression}
    \vspace{-3mm}
\end{figure}

\vspace{-1em}
\paragraph{Learning to Denoise}
We can optionally predict per-input point weights to make our solutions more robust to noise. We predict these via a fully connected network $w_j = \rho(\phi(\bm x_j; \mathcal{X}, \theta); \theta_w) \in \RR$ mapping per-point input features to weights. Instead of Eq.~\ref{eq:linsys_learned}, we then solve the \emph{weighted} ridge regression problem:
\begin{equation}
    \bm \alpha = (\bm W \bm G \bm W + \lambda \bm I)^{-1} \bm W \bm y
\end{equation}
where $\bm W = \text{diag}(w_1, \ldots w_s)$ is a diagonal matrix of per-input-point weights. Figure~\ref{fig:weighted_ridge_regression} shows the effect of weighted versus unweighted ridge regression in the presence of noise on a toy example.

\vspace{-0.5em}
\section{Experiments}\label{sec:experiments}
We first evaluate the effectiveness of \LongNames on the tasks of single object reconstruction (Section~\ref{sec:reconstruction_shapenet}) and partial object completion (Section~\ref{sec:completion}) using the ShapeNet~\cite{chang2015shapenet} dataset. 
Next, we highlight \ShortName's ability to generalize by evaluating the tasks of out-of-category shape generlization (Section~\ref{sec:generalization}), generlization to full scenes (Section~\ref{sec:reconstruction_scannet}), and generlization to different sampling densities (Section~\ref{sec:convergence}). Finally, in Section~\ref{sec:ablations}, we ablate the design choices for our backbone architecture. 
\input{figures/shapenet_reconstruction}
\begin{table*}[t]
    \setlength{\tabcolsep}{6pt}
    \renewcommand{\arraystretch}{1.1}
	\centering
	\resizebox{\textwidth}{!}{
    \begin{tabular}{l|cccccc|cccccc|cccccc}
			\toprule
			& \multicolumn{6}{c|}{Noise free} & \multicolumn{6}{c|}{Noise std. = 0.0025} & \multicolumn{6}{c}{Noise std. = 0.005} \\
			\hline
			& \multicolumn{2}{c}{IoU~$\uparrow$} & \multicolumn{2}{c}{Chamfer~$\downarrow$} & \multicolumn{2}{c|}{Normal C.~$\uparrow$}
			& \multicolumn{2}{c}{IoU~$\uparrow$} & \multicolumn{2}{c}{Chamfer~$\downarrow$} & \multicolumn{2}{c|}{Normal C.~$\uparrow$}  
			& \multicolumn{2}{c}{IoU~$\uparrow$} & \multicolumn{2}{c}{Chamfer~$\downarrow$} & \multicolumn{2}{c}{Normal C.~$\uparrow$} \\
            & mean & std. & mean & std. & mean & std. 
            & mean & std. & mean & std. & mean & std.
            & mean & std. & mean & std. & mean & std. \\
            \hline
 			SPSR~\cite{kazhdan2013screened} & 0.772 & 0.162 & 0.122 & 0.069 & 0.847 & 0.061
 			                                & 0.759 & 0.163 & 0.125 & 0.066 & 0.847 & 0.060
 			                                & 0.735 & 0.169 & 0.133 & 0.067 & 0.843 & 0.060 \\
 			OccNet~\cite{mescheder2019occnet} & 0.773 & 0.162 & 0.068 & 0.048 & 0.902 & 0.073 
 			                                  & 0.771 & 0.164 & 0.069 & 0.051 & 0.903 & 0.072 
 			                                  & 0.699 & 0.172 & 0.192 & 0.137 & 0.888 & 0.074 \\
 			C-OccNet~\cite{peng2020convoccnet} & 0.810 & 0.116 & 0.051 & 0.018 & 0.922 & 0.052 
 			                                   & 0.820 & 0.112 & 0.049 & 0.019 & 0.924 & 0.051
 			                                   & 0.866 & 0.089 & 0.080 & 0.040 & 0.937 & 0.044 \\
 			C-OccNet*~\cite{peng2020convoccnet} & 0.823 & 0.105 & 0.048 & 0.016 & 0.928 & 0.048
 			                                    & 0.847 & 0.094 & 0.043 & 0.015 & 0.932 & 0.046
 			                                    & 0.863 & 0.088 & 0.078 & 0.031 & 0.937 & 0.045 \\
 			NS~\cite{williams2020nsplines} & 0.864 & 0.151 & 0.051 & 0.071 & 0.926 & 0.059
 			                               & 0.831 & 0.147 & 0.054 & 0.064 & 0.919 & 0.057
 			                               & 0.791 & 0.155 & 0.121 & 0.167 & 0.900 & 0.055 \\
 			\textbf{Ours} & \textbf{0.949} & 0.053 & \textbf{0.024} & 0.010 & \textbf{0.954} & 0.042 
 			              & \textbf{0.914} & 0.061 & \textbf{0.028} &0.010 & \textbf{0.947} & 0.043
 			              & \textbf{0.883} & 0.074 & \textbf{0.066} & 0.018 & \textbf{0.939} & 0.041 \\
			\bottomrule
			
	\end{tabular}
	}
 	\vspace{-1.5mm}
	\caption{\textbf{Single object  reconstruction on ShapeNet~\cite{chang2015shapenet}}. \ShortName consistently outperforms strong baselines on standard metrics: IoU, Chamfer distance, and Normal Consistency, across all 13 categories.}
	\label{tab:reconstruction}
	\vspace{-4mm}
\end{table*}

\parahead{Baselines} For ShapeNet reconstruction, we compare our method to OccNet~\cite{mescheder2019occnet}, Conv-OccNet~\cite{peng2020convoccnet},  SPSR~\cite{kazhdan2013screened}, and Neural Splines~\cite{williams2020nsplines}. On the task of completion, we compare against Conv-OccNet~\cite{peng2020convoccnet}. For out-of-distribution shape reconstruction, we compare with OccNet~\cite{mescheder2019occnet}, Conv-OccNet~\cite{peng2020convoccnet}, LIG~\cite{jiang2020lig}, and Neural Splines~\cite{williams2020nsplines}, while on the task of full scene reconstruction we use Conv-OccNet~\cite{peng2020convoccnet}, SPSR~\cite{kazhdan2013screened}, and NS~\cite{williams2020nsplines} as baselines. Combined, these methods cover a broad spectrum of 3D shape reconstruction approaches and represent SoTA in their respective categories depicted in Fig.~\ref{fig:related_work}.

\parahead{Metrics}
We use 3 metrics for quantitative evaluation: 
\textit{Intersection over Union (IoU)} is computed by sampling a set of 100k points in the volume around a watertight shape and computing the IoU of the set of inside points for the predicted and ground truth shapes. IoU indicates how well the predicted shape agrees with the ground truth both near and away from the surface.
\textit{L2 Chamfer Distance} is evaluated by sampling 100k points on the predicted and ground truth surfaces (extracted as meshes using marching cubes), then computing the average shortest distance between all pairs of points. Chamfer distance measures how accurately each method reconstructs the surface of the input shape.
\textit{Normal Correlation} is computed as the average dot product between the normals at pairs of nearest points on the ground truth and predicted shapes and evaluates how well each method does at preserving the surface direction. We use the same 100k samples as for Chamfer distance to compute this metric.

\vspace{-0.25em}
\subsection{Single Object Reconstruction on ShapeNet}\label{sec:reconstruction_shapenet} 
We evaluate \ShortName's performance against strong baselines in reconstructing objects from 13 categories of the ShapeNet dataset. As input to all methods we use 1000 randomly sampled surface points to which we add Gaussian noise of different magnitudes. For learning based methods (Conv-Occnet, OccNet, Ours), we train a single model across all 13 categories per noise level. Since both \ShortName and Neural Splines utilize pairs of points spread along the normals, we train a version of Conv-OccNet with (C-OccNet*) and without (C-OccNet) these points. Table~\ref{tab:reconstruction} shows that \ShortName achieves large improvements across all metrics, reaching near 95\% IoU on noise-free reconstruction. Figure~\ref{fig:shapenet_reconstruction}, which shows reconstructions at the middle noise level, clearly demonstrates how \ShortName recovers fine details like the cars' side-view mirror, the cord on the lamp, and the bulges on the chair legs. In the supplemental, we provide per-category results, additional figures, and ablations on different numbers of input points. 

\vspace{-0.25em}
\subsection{Shape Completion on ShapeNet}\label{sec:completion} 
Albeit using input points as anchors, thanks to the global support of the kernel, \ShortName can learn to recover an entire shape from partial input. To demonstrate that, we sample a point cloud from up to 50~\% of a shape surface along one of the principal axes, and supervise \ShortName to predict the full shape. We train a separate model per shape category for each of 13 ShapeNet categories. Table~\ref{tab:shapenet_completion} presents quantitative results across all categories for this task. \ShortName achieves on-par Chamfer and Normal correlation as C-OccNet with substantially better IoU. The top row of Fig.~\ref{fig:motivation} shows an example of completing a truck shape from very partial input. Note how \ShortName learned to leverage shape symmetry to faithfully recover unobserved regions like the wheels. The appendix shows per-category quantitative and qualitative results. %
\begin{table}[t]
    \setlength{\tabcolsep}{6pt}
    \renewcommand{\arraystretch}{1.1}
	\centering
	\resizebox{\columnwidth}{!}{
    \begin{tabular}{l|cccccc}
			\toprule
			& \multicolumn{2}{c}{IoU~$\uparrow$} & \multicolumn{2}{c}{Chamfer~$\downarrow$} & \multicolumn{2}{c}{Normal C.~$\uparrow$} \\ %
            & mean & std. & mean & std. & mean & std.\\ %
            \hline
 			C-OccNet~\cite{peng2020convoccnet} & 0.770 & 0.152 & \textbf{0.075} & 0.068 & \textbf{0.909} & 0.059 \\ %
 			\textbf{Ours} & \textbf{0.819} & 0.171 & 0.077 & 0.091 & 0.907 & 0.067 \\%  & \textbf{0.875} & 0.130  \\
			\bottomrule
			
	\end{tabular}
	}
	\caption{\textbf{Object completion from partial point clouds.}}
	\label{tab:shapenet_completion}
\end{table}

\subsection{Out of Category Generalization}\label{sec:generalization}
Generalization to categories beyond the train set is key to making learnable methods useful in the wild. To evaluate \ShortName on this task we train all methods on 6 of the ShapeNet categories (\textit{airplane, lamp, display, rifle, chair, cabinet}) and evaluate on the other 7 (\textit{bench, car, loudspeaker, sofa, table, telephone, watercraft}). Table~\ref{tab:generalization} presents quantitative statistics for this task using the standard metrics. \ShortName greatly outperforms both learned and non-learned baselines. Furthermore, we note in brackets the decrease in performance compared to the model trained on all categories. \ShortName, with a minimal $1.1\%$ drop in IoU, aligns with data-free methods thanks to its test-time adaptation ability. We point out that LIG only provides models pretrained on all categories, which sets an upper bound on its generalization performance. The distinct differences between \ShortName and baselines are readily apparent in Figure~\ref{fig:shapenet_generalization}.
\input{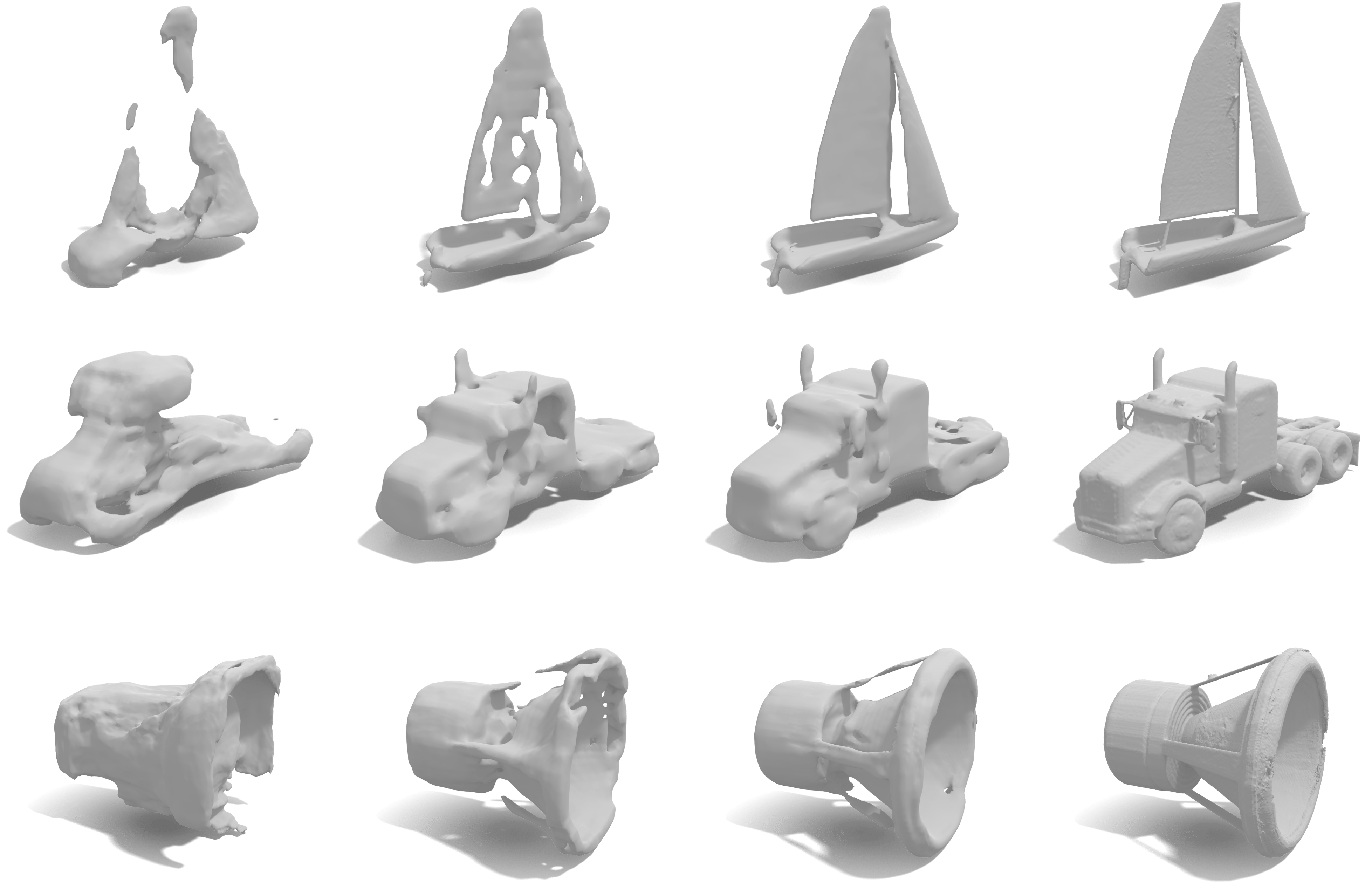}
\vspace{-3mm}
\definecolor{darkgreen}{RGB}{0,150,0}
\begin{table}[t]
    \setlength{\tabcolsep}{6pt}
    \renewcommand{\arraystretch}{1.1}
	\centering
	\resizebox{\columnwidth}{!}{
    \begin{tabular}{lccccc}
			\toprule
			& \multicolumn{1}{c}{IoU~$\uparrow$} & \multicolumn{1}{c}{Chamfer~$\downarrow$} & \multicolumn{1}{c}{Normal C.~$\uparrow$} \\
            \midrule
 			OccNet~\cite{mescheder2019occnet} & 0.603~(\textcolor{red}{-20.4\%}) & 0.134~(\textcolor{red}{0.070}) & 0.829~(\textcolor{red}{-8.3\%}) \\
 			C-OccNet~\cite{peng2020convoccnet} & 0.734~(\textcolor{red}{-9.5\%}) & 0.074~(\textcolor{red}{0.023}) & 0.895~(\textcolor{red}{-2.9\%}) \\
 			C-OccNet*~\cite{peng2020convoccnet} & 0.785~(\textcolor{red}{-4.9\%}) & 0.064~(\textcolor{red}{0.013}) & 0.911~(\textcolor{red}{-1.7\%})\\
 		    LIG~\cite{jiang2020lig} & 0.518~(N.A.) & 0.112~(N.A.) & 0.536~(N.A.) \\
 			NS~\cite{williams2020nsplines} & 0.869~(0.0\%) & 0.049~(0.000) & 0.924~(0.0\%) \\
 			\textbf{Ours} & \textbf{0.938~(\textcolor{red}{-1.1\%})} & \textbf{0.028~(\textcolor{red}{0.003})}  & \textbf{0.939~(\textcolor{red}{-1.0\%})} \\
			\bottomrule
	\end{tabular}
	}
	\caption{\textbf{Generalization capacity of object-level 3D reconstruction from sparse points clouds.} We train all models using 6 ShapeNet categories (\textit{airplane, lamp, display, rifle, chair, cabinet}) and evaluate them on the remaining 7 (\textit{bench, car, loudspeaker, sofa, table, telephone, watercraft}). The numbers in the brackets denote the difference in performance with the model trained on all categories.}
	\label{tab:generalization}
\end{table}
\input{figures/scannet_reconstruction}
\begin{table}[t]
    \setlength{\tabcolsep}{9pt}
    \renewcommand{\arraystretch}{1.1}
	\centering
	\resizebox{\columnwidth}{!}{
    \begin{tabular}{lcccccc}
			\toprule
			& \multicolumn{1}{c}{Chamfer~$\downarrow$} & \multicolumn{1}{c}{Normal C.~$\uparrow$} \\
            \midrule
 			C-OccNet (w. walls)~\cite{peng2020convoccnet} & 0.133 & 0.779  \\
 			C-OccNet (w.o. walls)~\cite{peng2020convoccnet} & 0.074 & 0.843 \\
 			SPSR~\cite{kazhdan2013screened} & 0.060 & 0.871 \\
 			NS \cite{williams2020nsplines} & 0.060 & \textbf{0.876} \\
 			\textbf{Ours} & \textbf{0.032} & 0.873  \\
			\bottomrule
	\end{tabular}
	}
	\caption{\textbf{Scene-level 3D reconstruction from sparse point clouds on ScanNet~\cite{dai2017scannet}.} All methods use 10~000 input points for each scene.}
	\label{tab:scannet_recon}
	\vspace{-2mm}
\end{table}

\vspace{2mm}
\subsection{Scene Reconstruction on ScanNet}\label{sec:reconstruction_scannet}
Next, we extend beyond single objects and evaluate \ShortNames on ScanNet scenes. For this experiment, we followed the setup in \cite{peng2020convoccnet} and trained our model on synthetic scenes consisting of random ShapeNet object placements. We found the synthetic floors and walls, added by \cite{peng2020convoccnet} to the training set, harmed performance and, hence, trained our method without them. We report C-OccNet's results with and without walls for completeness. According to Table~\ref{tab:scannet_recon}, for 10K input points, \ShortName achieves an average Chamfer distance of about half of the next best method. Figure~\ref{fig:scannet_reconstruction} shows a comparison to baselines on 2 reconstructed rooms. Now how our method better captures small details such as the stepladder and shelf.

\subsection{Point Density Generalization}\label{sec:convergence}
\begin{figure}
    \centering
    \includegraphics[width=\columnwidth]{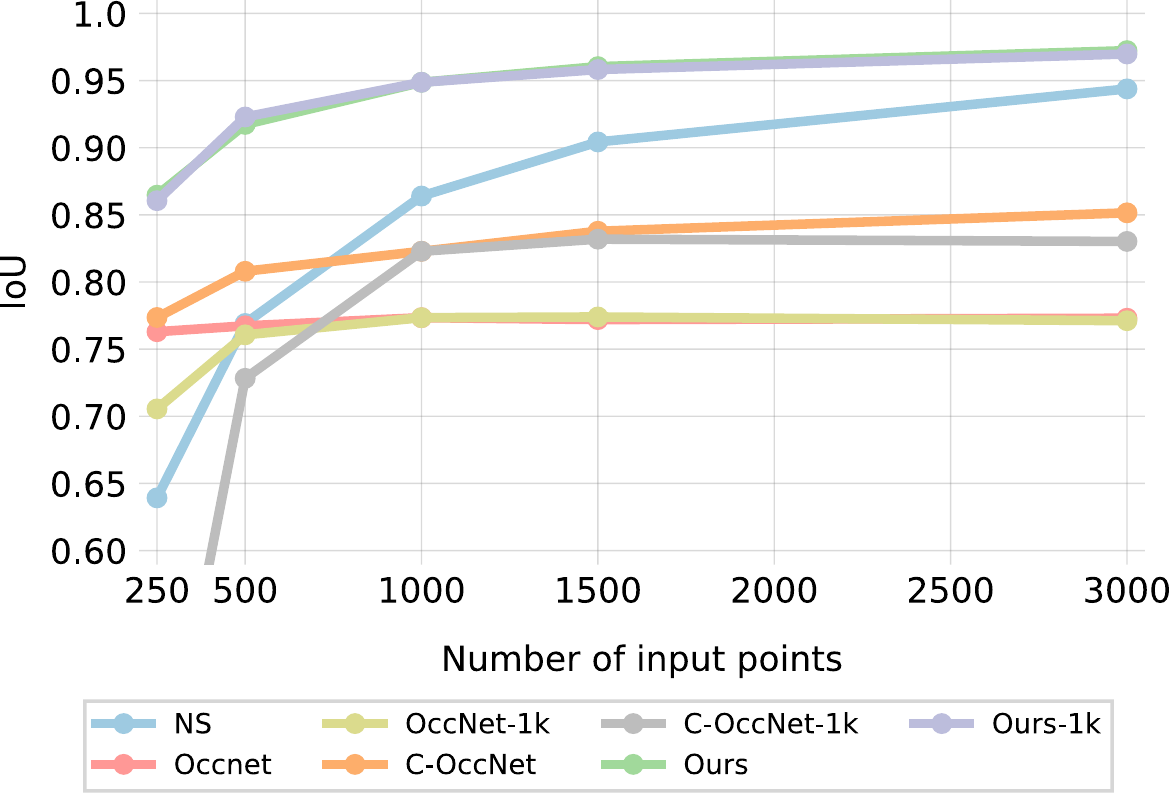}
    \caption{\textbf{ShapeNet IoU vs. number of input points.} Curves ending in "-1k" correspond to methods trained on 1000 points, and other methods were trained and evaluated on the same number of points. Our method performs well in the sparse and dense regimes and does not decay when trained and tested on different point densities.}
    \label{fig:iou_vs_points}
    \vspace{-5mm}
\end{figure}

In real-world applications, point density may differ between train and test times. A good data-driven prior should compensate for lack of data (i.e. sparse inputs) without hindering data-rich settings (i.e. dense inputs). Therefore, we evaluate the response of \ShortName and various baseline methods to changes in input sampling density. We trained each method on 1000 input points and evaluated it on varying numbers of input samples (between 250 and 3000). To report the upper-bound performance of each method, we train additional models on each density value. Figure~\ref{fig:iou_vs_points} shows the mean IoU of each method versus the number of input points. Curves with labels ending in "-1k" were trained on 1000 points, and otherwise, were trained and tested on the same number of points. OccNet shows no response to increased sampling density (even at train time). Although C-OccNet marginally improves when trained on denser data, it does not improve when evaluated with more points than it was trained on. 
The performance of Neural Splines improves for denser inputs, but is poor on sparse inputs as expected from data-free methods. Finally, our method works well in sparse settings and improves with increasing density. Moreover, it does not degrade if trained and tested on different sampling densities (the gray and green curves are nearly identical). 

\subsection{Ablations}\label{sec:ablations}
We conduct an ablation study of our design choices on the task of shape reconstruction on ShapeNet. We experiment with using different per-point feature dimensions and whether to include the surface $L_1$ loss, $L(f)_{L1}$. Table~\ref{tab:shaoenet_ablation} summarizes the results.
\begin{table}[t]
    \setlength{\tabcolsep}{10pt}
	\centering
	\resizebox{\columnwidth}{!}{
    \begin{tabular}{l|cccc}
			\toprule
			& \multicolumn{4}{c}{feature dimension} \\
			& 8 & 16 & 32 & 64  \\
            \midrule
 			without $L(f)_{L1}$  & 0.939 & 0.941 & 0.942 & 0.942 \\
 			with  $L(f)_{L1}$  & 0.945 & 0.947 & \textbf{0.949} & \textbf{0.949} \\
			\bottomrule
	\end{tabular}
	}
	\vspace{0.1mm}
	\caption{\textbf{Ablation study}~(Section~\ref{sec:reconstruction_shapenet}). \ShortNames benefits from the L1 surface loss and work well even with small feature dimensions. %
	Values in the table are mean IoU on the test set.}
	\label{tab:shaoenet_ablation}
	\vspace{-3mm}
\end{table}

\section{Conclusion and Limitations}
We presented a novel method for reconstructing and completing 3D shapes from sparse point clouds. Our method outperforms the state-of-the-art on object reconstruction and completion as well as scene reconstruction, while demonstrating strong generalization capability (both with respect to shape categories and input sampling density). While our method pushes the boundary on many fronts, it still has several limitations which we plan to address in future work: First, our current kernel implementation requires a dense linear solve, which limits the number of evaluation points to around 12k on a V100 GPU. State-of-the-art Kernel solvers in the literature (e.g. \cite{rudi2018falkon}) have scaled up to millions of points by leveraging techniques such as Nystr{\"o}m sampling. We plan to investigate how to leverage these approaches to handle larger inputs. Furthermore, we would like to investigate kernels with spatial decay to sparsify our linear system and scale our method to very large inputs. A second limitation is the requirement of oriented points. While these are usually available from sensors, they can be noisy. Thus, in the future we would like to incorporate normal prediction into our method so it can operate on unoriented point clouds.

\newpage
{\small
\typeout{}
\bibliographystyle{ieee_fullname}
\bibliography{main.bib}
}

\newpage
\appendix
\onecolumn
\section{Neural Spline Kernel Equation}
The Neural Spline \cite{williams2020nsplines} kernel is defined as the limiting kernel for an infinitely wide ReLU network with either Gaussian or Uniform initialization (using Kaiming-He \cite{he2015delving} initialization). In our implementation we use the Gaussian initialized version which has the following closed form solution:
\begin{equation}
    K_\text{NS}(\bm x, \bm x') = \frac{\|\tilde{\bm x}\| \|\tilde{\bm x}'\|}{\pi} \big(\sin\theta + 2 (\pi - \theta) \cos\theta \big) \qquad \theta = \measuredangle(\bm x, \bm x')
\end{equation}
where $\tilde{\bm x} = (\bm x, 1), \tilde{\bm x}' = (\bm x', 1)$ are the vectors $\bm x$ and $\bm x'$ expressed in homogeneous coordinates and $\theta = \measuredangle(\tilde{\bm x}, \tilde{\bm x}')$ is the angle between the input vectors in homogeneous coordinates. In practice we compute the angle using the formula from Kahan~\cite{kahancomputing}:

\begin{equation}
    \theta = 2 \arctan(\frac
    {\bigl\|\ \|\tilde{\bm x}'\| \tilde{\bm x} - \|\tilde{\bm x}\| \tilde{\bm x}'\ \bigr\|}
    {\bigl\|\ \|\tilde{\bm x}'\| \tilde{\bm x} + \|\tilde{\bm x}\| \tilde{\bm x}'\ \bigr\|}),
\end{equation}

which is numerically stable, especially with small angles.

\section{The Effect of Noise Filtering in 3D}
\begin{figure*}
  \centering
  \begin{tikzpicture}
    \node (fig) {
        \ifHighResImages
          \includegraphics[width=\linewidth]{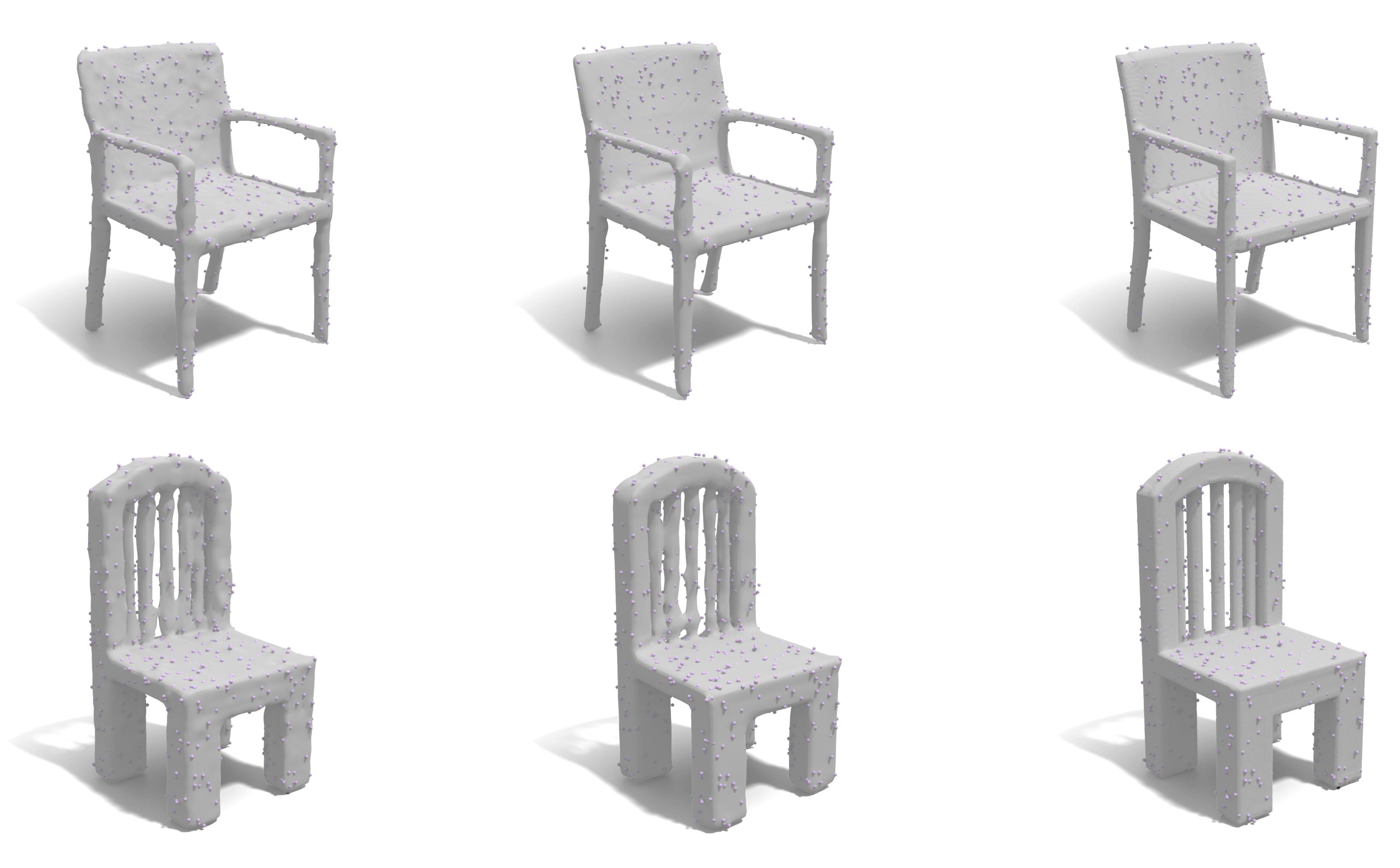}
        \else
          \includegraphics[width=\linewidth]{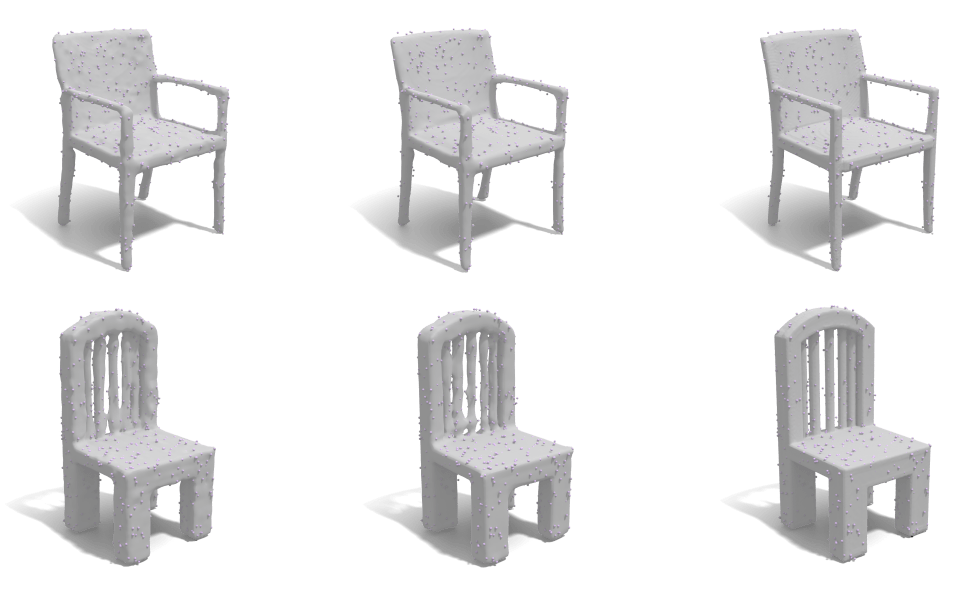}
        \fi
    };
    \node[anchor=west, text width=5cm] at (-7.75,-5.75) 
    {\small Ours (w/o weights)};
    \node[anchor=west, text width=3cm] at (-1.05,-5.75) 
    {\small Ours (weighted)};
    \node[anchor=west, text width=3cm] at (5.75,-5.75) 
    {\small Ground truth};
  \end{tikzpicture}
  \caption{\textbf{The effect of noise filtering versus regularization.} The left column shows reconstructions using our method without any noise filtering and $0.1$ regularization in the kernel ridge regression. The middle column shows these same models reconstructed with additional noise filtering (Section~\ref{sec:neural_kernel_fields}). Note how the regularized model still has bumps caused by the noisy input points while these are smoothed out by the filtering module.}   \label{fig:noise3d}
\end{figure*}

Figure~\ref{fig:noise3d} shows the effect of weighting (Section~\ref{sec:neural_kernel_fields}) to filter noise in the input points. The left column shows our reconstruction without these learned weights, the middle column shows the effect of adding weighting, while the right column shows the ground truth surface. Notice how the weighted model is smoother and does not interpolate the input noise.

\section{More Extreme Generalization}
Table~\ref{tab:generalization_chairs} and Figure~\ref{fig:extreme_generalization} compare our reconstruction results using a model trained \textit{only on chairs} to reconstruct the other 12 ShapeNet categories (airplane, bench, cabinet, car, display, lamp, loudspeaker, rifle, sofa, table, telephone, watercraft) against a model trained on all categories. The experimental setup is identical to Section~\ref{sec:generalization} (1000 input points) except the model is trained only on chairs. Note how the performance of model trained only on chairs only drops slightly compared to the model trained on all categories. 
\definecolor{darkgreen}{RGB}{0,150,0}
\begin{table}[t]
    \setlength{\tabcolsep}{6pt}
    \renewcommand{\arraystretch}{1.1}
	\centering
    \begin{tabular}{l|c|c|c||c|c|c}
			\toprule
			& \multicolumn{3}{c||}{Pretrain on Chairs} & \multicolumn{3}{c}{Pretrain on All}\\
			\hline
			& \multicolumn{1}{c|}{IoU~$\uparrow$} & \multicolumn{1}{c|}{Chamfer~$\downarrow$}& \multicolumn{1}{c||}{Normal C.~$\uparrow$} & \multicolumn{1}{c|}{IoU~$\uparrow$} & \multicolumn{1}{c|}{Chamfer~$\downarrow$}& \multicolumn{1}{c}{Normal C.~$\uparrow$}\\
			\hline
			airplane &     0.922 & 0.021 & 0.945 & 0.951 & 0.016 & 0.962 \\
			bench &        0.898 & 0.024 & 0.936 & 0.908 & 0.022 & 0.940 \\
			cabinet &      0.938 & 0.043 & 0.947 & 0.968 & 0.028 & 0.962 \\
			car &          0.913 & 0.037 & 0.882 & 0.937 & 0.030 & 0.913 \\
			chair &        0.946 & 0.026 & 0.962 & 0.943 & 0.027 & 0.960 \\
			display &      0.967 & 0.028 & 0.971 & 0.976 & 0.023 & 0.978 \\
			lamp &         0.895 & 0.040 & 0.928 & 0.920 & 0.024 & 0.940 \\
			loudspeaker &  0.931 & 0.059 & 0.935 & 0.965 & 0.033 & 0.952 \\
			rifle &        0.889 & 0.115 & 0.937 & 0.957 & 0.012 & 0.970 \\
			sofa &         0.971 & 0.025 & 0.967 & 0.974 & 0.024 & 0.969 \\
			table &        0.939 & 0.028 & 0.964 & 0.951 & 0.025 & 0.969 \\
			telephone &    0.985 & 0.018 & 0.986 & 0.988 & 0.017 & 0.988 \\
			watercraft&    0.936 & 0.039 & 0.934 & 0.955 & 0.019 & 0.950 \\
			\hline
			mean&          0.929 & 0.036 & 0.939 & 0.949 & 0.024 & 0.954 \\ 
			\bottomrule
	\end{tabular}
	\caption{Comparison between model trained only on chairs (left column) to model trained on all categories.}
	\label{tab:generalization_chairs}
\end{table}
\input{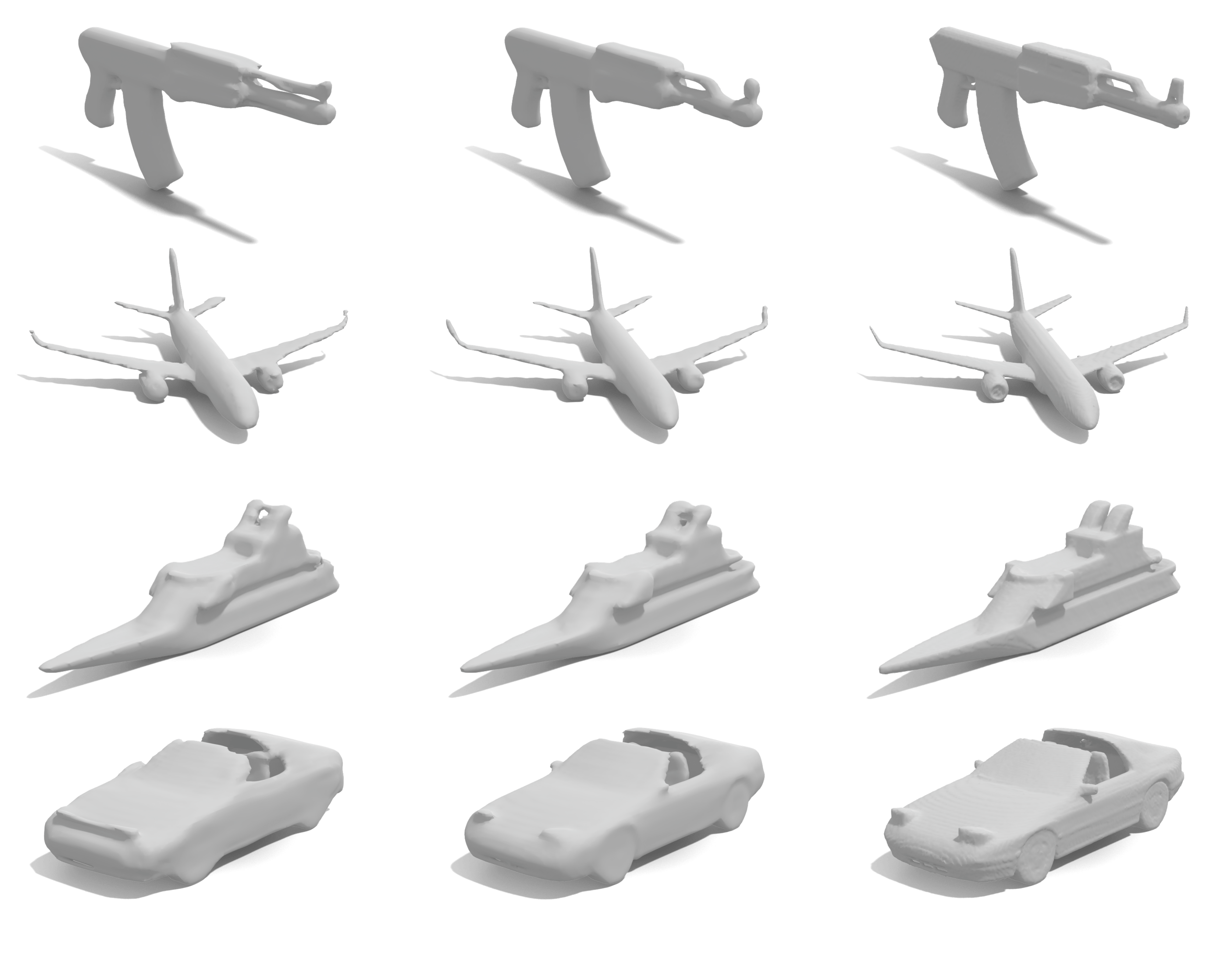}

\section{Inference Timings}
Our method uses a convex test-time optimization to perform inference of 3D shapes. We report the timing of each part of our method for the ShapeNet reconstruction (Section~\ref{sec:reconstruction_shapenet}) and ScanNet reconstruction (Section~\ref{sec:reconstruction_scannet}) experiments in Table~\ref{tab:timings}. With $1000$ input points for ShapeNet, we evaluated on a grid of size $128^3$ ($2.1$M points), and with $8000$ input points for ScanNet, we evaluated on a grid of size $256^3$ ($16.78$M points. We implemented the kernel evaluation as a single monolithic CUDA kernel and report the timings on a Quadro GV100 GPU.

\begin{table}[t]
\centering
\begin{tabular}{@{}c| c | c}
\toprule
& ShapeNet & ScanNet \\
\hline
Encoder & 12.9ms  & 229.8ms \\
Decoder & 0.3ms & 0.42ms \\
Solve & 30.3ms & 3142ms \\
Eval & 193.5ms & 13254ms \\
\bottomrule
\end{tabular}
\caption{Timings on ShapeNet (1k input points and 2.1 million evaluation points) and ScanNet (10k input points and 16.9 million eval points). }
\label{tab:timings}
\end{table}

\section{Additional ShapeNet Reconstruction Figures}
Figure~\ref{fig:shapenet_reconstruction_supp} shows additional reconstruction comparisons (with 0.0025) noise as described in Section~\ref{sec:reconstruction_shapenet}.
\input{figures/shapenet_reconstruction_supplementary}

\section{Additional ShapeNet Generalization Figures}
Figure~\ref{fig:shapenet_completion_supp} shows additional reconstructions for the out-of-category reconstruction experiment described in Section~\ref{sec:generalization}.

\section{Additional Completion Figures}
Figure~\ref{fig:shapenet_completion_supp} shows additional completion comparisons for the experiment described in Section~\ref{sec:completion}.
\input{figures/shapenet_completion_supplementary}

\section{Per-Category ShapeNet Results}
Tables \ref{tab:category_level_reconstruction} and \ref{tab:category_level_completion} report the per-category reconstruction and completion results respectively for the experiments described in Sections~\ref{sec:reconstruction_shapenet} and ~\ref{sec:completion}.
\begin{table*}[t]
    \setlength{\tabcolsep}{6pt}
    \renewcommand{\arraystretch}{1.1}
	\centering
	\resizebox{\textwidth}{!}{
    \begin{tabular}{l|cccc|cccc|cccc}
			\toprule
			\multicolumn{13}{c}{Noise-Free} \\
			\hline
			& \multicolumn{4}{c|}{IoU~$\uparrow$} & \multicolumn{4}{c|}{Chamfer~$\downarrow$}& \multicolumn{4}{c}{Normal C.~$\uparrow$} \\
			& OccNet & C-OccNet* & NS & Ours & OccNet & C-OccNet* & NS & Ours & OccNet & C-OccNet* & NS & Ours \\
			\hline
			airplane &    0.752 & 0.811 & 0.775 & \textbf{0.951} & 0.054 & 0.036 & 0.103 & \textbf{0.016} & 0.900 & 0.927 & 0.898 & \textbf{0.962} \\
			bench &       0.713 & 0.723 & 0.768 & \textbf{0.908} & 0.052 & 0.045 & 0.065 & \textbf{0.022} & 0.889 & 0.900 & 0.901 & \textbf{0.940} \\
			cabinet &     0.869 & 0.898 & 0.921 & \textbf{0.968} & 0.060 & 0.049 & 0.041 & \textbf{0.028} & 0.931 & 0.950 & 0.939 & \textbf{0.962} \\
			car &         0.841 & 0.873 & 0.911 & \textbf{0.937} & 0.069 & 0.051 & 0.037 & \textbf{0.030} & 0.896 & 0.898 & 0.903 & \textbf{0.913} \\
			chair &       0.740 & 0.811 & 0.858 & \textbf{0.943} & 0.076 & 0.051 & 0.045 & \textbf{0.027} & 0.896 & 0.933 & 0.933 & \textbf{0.960} \\
			display &     0.825 & 0.854 & 0.938 & \textbf{0.976} & 0.062 & 0.048 & 0.030 & \textbf{0.023} & 0.932 & 0.960 & 0.964 & \textbf{0.978} \\
			lamp &        0.550 & 0.751 & 0.834 & \textbf{0.920} & 0.144 & 0.058 & 0.047 & \textbf{0.024} & 0.819 & 0.902 & 0.915 & \textbf{0.940} \\
			loudspeaker & 0.833 & 0.892 & 0.938 & \textbf{0.965} & 0.090 & 0.059 & 0.041 & \textbf{0.033} & 0.910 & 0.938 & 0.945 & \textbf{0.952} \\
			rifle &       0.678 & 0.757 & 0.936 & \textbf{0.957} & 0.057 & 0.038 & 0.021 & \textbf{0.012} & 0.860 & 0.915 & 0.960 & \textbf{0.970} \\
			sofa &        0.876 & 0.893 & 0.927 & \textbf{0.974} & 0.055 & 0.047 & 0.041 & \textbf{0.024} & 0.939 & 0.952 & 0.949 & \textbf{0.969} \\
			table &       0.768 & 0.785 & 0.801 & \textbf{0.951} & 0.059 & 0.048 & 0.065 & \textbf{0.025} & 0.923 & 0.948 & 0.926 & \textbf{0.969} \\
			telephone &   0.915 & 0.904 & 0.969 & \textbf{0.988} & 0.035 & 0.035 & 0.021 & \textbf{0.017} & 0.973 & 0.979 & 0.983 & \textbf{0.988} \\
			watercraft&   0.737 & 0.825 & 0.894 & \textbf{0.955} & 0.083 & 0.046 & 0.044 & \textbf{0.019} & 0.870 & 0.909 & 0.930 & \textbf{0.950} \\
			\hline
			mean&   0.773 & 0.823 & 0.864 & \textbf{0.949} & 0.068 & 0.048 & 0.051 & \textbf{0.024} & 0.902 & 0.928 & 0.926 & \textbf{0.954} \\ 
			\bottomrule
			
			\toprule
			\multicolumn{13}{c}{0.0025 Noise} \\
			\hline
			& \multicolumn{4}{c|}{IoU~$\uparrow$} & \multicolumn{4}{c|}{Chamfer~$\downarrow$}& \multicolumn{4}{c}{Normal C.~$\uparrow$} \\
			& OccNet& C-OccNet* & NS & Ours & OccNet & C-OccNet* & NS & Ours & OccNet & C-OccNet* & NS & Ours \\
			\hline
			airplane &    0.739 & 0.825 & 0.729 & \textbf{0.905} & 0.057 & 0.034 & 0.103 & \textbf{0.020} & 0.904 & 0.928 & 0.888 & \textbf{0.953} \\
			bench &       0.713 & 0.758 & 0.723 & \textbf{0.867} & 0.053 & 0.040 & 0.068 & \textbf{0.025} & 0.889 & 0.906 & 0.892 & \textbf{0.935} \\
			cabinet &     0.871 & 0.916 & 0.905 & \textbf{0.952} & 0.061 & 0.044 & 0.045 & \textbf{0.031} & 0.933 & 0.953 & 0.934 & \textbf{0.959} \\
			car &         0.839 & 0.877 & 0.892 & \textbf{0.921} & 0.068 & 0.052 & 0.041 & \textbf{0.033} & 0.895 & 0.902 & 0.896 & \textbf{0.911} \\
			chair &       0.740 & 0.837 & 0.825 & \textbf{0.912} & 0.077 & 0.045 & 0.050 & \textbf{0.030} & 0.896 & 0.937 & 0.926 & \textbf{0.956} \\
			display &     0.818 & 0.890 & 0.902 & \textbf{0.953} & 0.063 & 0.039 & 0.036 & \textbf{0.026} & 0.932 & 0.963 & 0.958 & \textbf{0.975} \\
			lamp &        0.547 & 0.774 & 0.784 & \textbf{0.880} & 0.153 & 0.050 & 0.053 & \textbf{0.026} & 0.824 & 0.907 & 0.906 & \textbf{0.936} \\
			loudspeaker & 0.829 & 0.910 & 0.922 & \textbf{0.952} & 0.091 & 0.052 & 0.046 & \textbf{0.035} & 0.912 & 0.943 & 0.940 & \textbf{0.952} \\
			rifle &       0.678 & 0.783 & 0.860 & \textbf{0.904} & 0.058 & 0.033 & 0.023 & \textbf{0.016} & 0.865 & 0.919 & 0.947 & \textbf{0.960} \\
			sofa &        0.879 & 0.913 & 0.905 & \textbf{0.956} & 0.055 & 0.041 & 0.047 & \textbf{0.028} & 0.937 & 0.956 & 0.942 & \textbf{0.966} \\
			table &       0.768 & 0.832 & 0.772 & \textbf{0.917} & 0.059 & 0.040 & 0.065 & \textbf{0.028} & 0.924 & 0.953 & 0.922 & \textbf{0.966} \\
			telephone &   0.909 & 0.931 & 0.932 & \textbf{0.969} & 0.036 & 0.029 & 0.027 & \textbf{0.020} & 0.973 & 0.980 & 0.975 & \textbf{0.986} \\
			watercraft&   0.732 & 0.843 & 0.857 & \textbf{0.926} & 0.086 & 0.041 & 0.050 & \textbf{0.022} & 0.874 & 0.913 & 0.918 & \textbf{0.945} \\
			\hline
			mean &   0.771 & 0.847 & 0.831 & \textbf{0.919} &  0.069 & 0.043 & 0.054 & \textbf{0.027} &  0.903 & 0.932 & 0.919 & \textbf{0.945} \\ 
			\bottomrule
			
			\toprule
			\multicolumn{13}{c}{0.005 Noise} \\
			\hline
			& \multicolumn{4}{c|}{IoU~$\uparrow$} & \multicolumn{4}{c|}{Chamfer~$\downarrow$}& \multicolumn{4}{c}{Normal C.~$\uparrow$} \\
			& OccNet& C-OccNet* & NS & Ours & OccNet & C-OccNet* & NS & Ours & OccNet & C-OccNet* & NS & Ours \\
			\hline
			airplane &    0.675 & 0.839 & 0.758 & \textbf{0.852} & 0.155 & 0.062 & 0.098 & \textbf{0.053} & 0.890 & 0.933 & 0.886 & \textbf{0.937} \\
			bench &       0.589 & 0.779 & 0.673 & \textbf{0.813} & 0.160 & 0.073 & 0.161 & \textbf{0.062} & 0.860 & 0.911 & 0.876 & \textbf{0.922} \\
			cabinet &     0.802 & 0.928 & 0.881 & \textbf{0.936} & 0.181 & 0.078 & 0.105 & \textbf{0.070} & 0.914 & 0.958 & 0.920 & \textbf{0.952} \\
			car &         0.804 & 0.888 & 0.869 & \textbf{0.899} & 0.182 & 0.095 & 0.095 & \textbf{0.077} & 0.891 & 0.905 & 0.879 & \textbf{0.902} \\
			chair &       0.652 & 0.859 & 0.779 & \textbf{0.876} & 0.217 & 0.081 & 0.119 & \textbf{0.071} & 0.884 & 0.944 & 0.910 & \textbf{0.946} \\
			display &     0.742 & 0.914 & 0.858 & \textbf{0.924} & 0.170 & 0.067 & 0.091 & \textbf{0.061} & 0.922 & 0.968 & 0.940 & \textbf{0.967} \\
			lamp &        0.478 & 0.796 & 0.701 & \textbf{0.827} & 0.421 & 0.099 & 0.171 & \textbf{0.065} & 0.802 & 0.914 & 0.868 & \textbf{0.921} \\
			loudspeaker & 0.785 & 0.924 & 0.900 & \textbf{0.937} & 0.236 & 0.091 & 0.108 & \textbf{0.080} & 0.899 & 0.947 & 0.925 & \textbf{0.946} \\
			rifle &       0.600 & 0.807 & 0.774 & \textbf{0.850} & 0.151 & 0.060 & 0.068 & \textbf{0.045} & 0.832 & 0.925 & 0.906 & \textbf{0.943} \\
			sofa &        0.818 & 0.929 & 0.889 & \textbf{0.936} & 0.159 & 0.072 & 0.095 & \textbf{0.065} & 0.925 & 0.961 & 0.931 & \textbf{0.957} \\
			table &       0.663 & 0.859 & 0.704 & \textbf{0.873} & 0.168 & 0.072 & 0.167 & \textbf{0.066} & 0.906 & 0.957 & 0.898 & \textbf{0.956} \\
			telephone &   0.847 & 0.944 & 0.892 & \textbf{0.945} & 0.107 & 0.050 & 0.072 & \textbf{0.049} & 0.966 & 0.982 & 0.958 & \textbf{0.980} \\
			watercraft &  0.695 & 0.863 & 0.808 & \textbf{0.890} & 0.216 & 0.074 & 0.147 & \textbf{0.056} & 0.861 & 0.921 & 0.890 & \textbf{0.931} \\
			\hline
			mean &        0.699 & 0.863 & 0.791 & \textbf{0.883} & 0.192 & 0.078 & 0.121 & \textbf{0.066} & 0.888 & 0.937 & 0.900 & \textbf{0.939} \\ 
			\bottomrule
	\end{tabular}
	}
	\caption{Per-category ShapeNet reconstruction results corresponding to the experiment described in Section~\ref{sec:reconstruction_shapenet}.}
	\label{tab:category_level_reconstruction}
\end{table*}

\begin{table*}[t]
    \setlength{\tabcolsep}{6pt}
    \renewcommand{\arraystretch}{1.1}
	\centering
    \begin{tabular}{l|cc|cc|cc|cc}
			\toprule
			& \multicolumn{2}{c|}{IoU~$\uparrow$} & \multicolumn{2}{c|}{Chamfer~$\downarrow$}& \multicolumn{2}{c|}{Normal C.~$\uparrow$} & \multicolumn{2}{c}{F-Score~$\uparrow$} \\
			& C-OccNet*& Ours & C-OccNet*& Ours & C-OccNet& Ours & C-OccNet*& Ours \\
			\hline
			airplane 	& 0.800 & 0.844 & 0.048 & 0.054 & 0.926 & 0.919 & 0.921 & 0.916 \\
			bench 		& 0.615 & 0.705 & 0.082 & 0.086 & 0.868 & 0.872 & 0.808 & 0.853 \\
			cabinet 	& 0.834 & 0.881 & 0.079 & 0.067 & 0.924 & 0.918 & 0.784 & 0.872 \\
			car 		& 0.862 & 0.891 & 0.059 & 0.047 & 0.899 & 0.899 & 0.859 & 0.912 \\
			chair 		& 0.731 & 0.790 & 0.092 & 0.091 & 0.906 & 0.910 & 0.805 & 0.854 \\
			display 	& 0.768 & 0.850 & 0.088 & 0.079 & 0.921 & 0.925 & 0.774 & 0.876 \\
			lamp 		& 0.620 & 0.685 & 0.138 & 0.159 & 0.864 & 0.866 & 0.751 & 0.797 \\
			loudspeaker & 0.808 & 0.851 & 0.101 & 0.105 & 0.904 & 0.902 & 0.701 & 0.814 \\
			rifle 		& 0.746 & 0.809 & 0.045 & 0.051 & 0.899 & 0.904 & 0.915 & 0.907 \\
			sofa 		& 0.837 & 0.864 & 0.073 & 0.075 & 0.923 & 0.916 & 0.823 & 0.866 \\
			table 		& 0.730 & 0.777 & 0.075 & 0.089 & 0.925 & 0.911 & 0.851 & 0.863 \\
			telephone 	& 0.886 & 0.906 & 0.046 & 0.048 & 0.964 & 0.958 & 0.920 & 0.922 \\
			watercraft 	& 0.761 & 0.830 & 0.067 & 0.061 & 0.887 & 0.912 & 0.829 & 0.884 \\
			\hline 
			mean & 0.770 & 0.819 & 0.075 & 0.077 & 0.909 & 0.907 & 0.837 & 0.875 \\
			\bottomrule
			
	\end{tabular}
	\caption{Per category completion results corresponding to the experiment described in Section~\ref{sec:completion}.}
	\label{tab:category_level_completion}
\end{table*}

\end{document}